%% file: arxiv.tex
\documentclass[10pt,twocolumn,letterpaper]{article}

\usepackage[pagenumbers]{cvpr} 


\definecolor{cvprblue}{rgb}{0.21,0.49,0.74}
\usepackage[pagebackref,breaklinks,colorlinks,allcolors=cvprblue]{hyperref}

\usepackage{tabularx}
\newcolumntype{C}{>{\centering\arraybackslash}X}
\usepackage{wrapfig}

\usepackage{ragged2e}
\usepackage{xspace}
\usepackage{makecell}

\title{Material Anything: Generating Materials for Any 3D Object via Diffusion}

\author{ Xin Huang\textsuperscript{1*}, Tengfei Wang\textsuperscript{2$^\dag$}, Ziwei Liu\textsuperscript{3}, Qing Wang\textsuperscript{1$^\dag$} \\
$^1$Northwestern Polytechnical University, $^2$Shanghai AI Lab, $^3$S-Lab, Nanyang Technological University \\
}

\begin{document}
\twocolumn[{%
\renewcommand\twocolumn[1][]{#1}%
\maketitle
\begin{center}
    \centering
    \vspace{-3mm}
    \captionsetup{type=figure}
    \includegraphics[width=0.98\hsize]{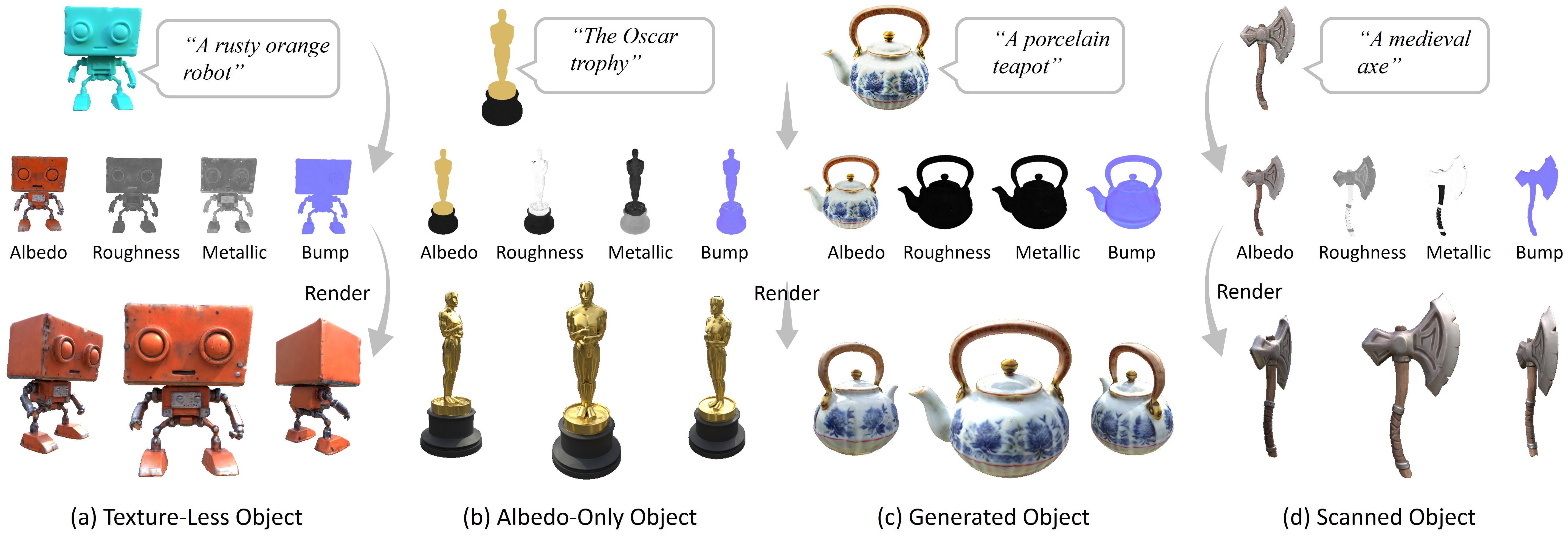}
    \vspace{-2mm}
    \caption{\textbf{Material Anything}: A feed-forward  PBR material generation model applicable to a diverse range of 3D meshes across varying texture and lighting conditions, including texture-less, albedo-only, generated, and scanned objects.
    }  
    \label{fig:teaser}
\end{center}%
}]

\let\thefootnote\relax\footnotetext{\textsuperscript{*} Work was done during an internship at Shanghai AI Lab.}
\let\thefootnote\relax\footnotetext{$^\dag$ Corresponding authors.}

\input{Sections/0_abstract}    
\input{Sections/1_intro}

\input{Sections/2_related}
\input{Sections/3_approach}
\input{Sections/4_exp}

\input{Sections/5_app}
\input{Sections/6_conclusion}

{
    \small
    \bibliographystyle{ieeenat_fullname}
    \bibliography{ref}
}
\input{Sections/X_suppl}

\end{document}

%% file: Sections/0_abstract.tex
\begin{abstract}
We present \textbf{Material Anything}, a fully-automated, unified diffusion framework designed to generate physically-based materials for 3D objects. Unlike existing methods that rely on complex pipelines or case-specific optimizations, Material Anything offers a robust, end-to-end solution adaptable to objects under diverse lighting conditions. Our approach leverages a pre-trained image diffusion model, enhanced with a triple-head architecture and rendering loss to improve stability and material quality. Additionally, we introduce confidence masks as a dynamic switcher within the diffusion model, enabling it to effectively handle both textured and texture-less objects across varying lighting conditions. By employing a progressive material generation strategy guided by these confidence masks, along with a UV-space material refiner, our method ensures consistent, UV-ready material outputs. Extensive experiments demonstrate our approach outperforms existing methods across a wide range of object categories and lighting conditions.
\end{abstract}
\vspace{-3mm}

%% file: Sections/1_intro.tex
\vspace{-2mm}
\section{Introduction}\label{sec:introduction}
Physically Based Rendering (PBR) involves complex interactions 
between geometry, materials, and illumination. 
High-quality physical materials ensure that 3D objects appear consistent and realistic under various lighting conditions, crucial for applications such as video games, virtual reality, and film production.
Given the meshes in Fig.~\ref{fig:teaser}, a skilled artist can create realistic textures and materials using software like Blender~\cite{blender} and Substance 3D Painter~\cite{substance}. However, this creation process is tedious and time-consuming, requiring expertise in graphic design.
Despite recent advances in 3D texture painting~\cite{richardson2023texture,chen2023text2tex,liu2023text,zeng2024paint3d}, they often fail to accurately model materials that disentangle light and texture, resulting in baked-in shading effects like unwanted highlights and shadows.

Several recent works have emerged to tackle the challenge of generating materials for 3D objects; however, they remain largely impractical due to their complexity and specificity. These approaches either require specific optimizations for each case~\cite{zhang2024dreammat,xu2023matlaber} or rely on multi-modal models like GPT4-V~\cite{achiam2023gpt} to retrieve materials for different parts of an object~\cite{zhang2024mapa,fang2024make}. Consequently, such approaches face significant challenges: \textbf{(1) Limited scalability.} Each case requires specific parameter adjustments, hindering the end-to-end automation of the creation process. \textbf{(2) Compromised robustness.} Complex pipelines that involve multiple models (\emph{e.g.,} SAM~\cite{kirillov2023segment} and GPT for segmentation and assignment) may lead to system instability. \textbf{(3) Limited generalization.}  Existing methods are sensitive to lighting and struggle to handle a broad spectrum of scenarios, including realistic lighting (\emph{e.g.,}  real-world scans), unrealistic lighting (\emph{e.g.,}  generated textures), and absence of lighting (\emph{e.g.,}  albedo).
To tackle these challenges, we propose \textbf{{Material Anything}},  a \emph{fully automated, stable, and universal} generative model for physical materials. Our method accepts any 3D mesh as input and generates high-quality material maps through a two-stage pipeline: image-space material generation and UV-space material refinement.

Given a 3D object, the image-space material diffusion model aims to produce PBR materials for each view of it. 
Considering the limited availability of PBR data, we leverage a pre-trained image diffusion model and adapt it to material estimation using a novel triple-head architecture and rendering loss, which together help stabilize training and bridge the gap between natural images and material maps. Once trained, this model can automatically generate materials for the views rendered from general 3D objects without predefined categories or part-level masks. 

To enable the image-space material diffusion model to support 3D objects across various lighting scenarios, we introduce a confidence mask to indicate illumination certainty and propose a data augmentation strategy to simulate various lighting conditions during training. 
(1) For meshes with \emph{realistic lighting effects}, the confidence mask is set to a higher value, enabling the model to utilize illumination cues to predict the materials accurately. 
(2) For meshes with \emph{lighting-free} textures, the confidence is set to low, allowing the model to generate materials based on prompts and global semantic cues. 
(3) For generated objects and texture-less objects (we initially use a texture generation method to create coarse textures), their textures may exhibit \emph{unrealistic lighting effects} that deviate from physical laws, often resulting in exaggerated highlights and shadows. In such cases, the confidence mask is adaptively set to varying values, ensuring the model relies on local semantic to generate plausible materials, as the lighting cues are unreliable. 


While the image-space model can effectively generate materials for each single view, directly applying it to a 3D object can lead to appearance inconsistency across views. To boost multi-view consistency, we introduce a confidence-aware progressive material generation scheme that uses the confidence mask to prompt our diffusion model to produce materials consistent with previous views.  After progressively generating the materials for all views, we project them into UV space for further refinement, achieving 3D-consistent and high-quality UV maps that are user-friendly and easy to edit.

Together, these components enable {Material Anything} to achieve remarkable performance in material generation. To train the model, we build a \emph{Material3D dataset}, comprising over 80K objects with high-quality PBR materials and UV unwrapping. Extensive experiments demonstrate a significant improvement of our method over current approaches. We summarize our contributions as follows:
\begin{itemize}[noitemsep,topsep=0pt]
    \item A fully automated, stable, and universal model to generate physical materials for diverse 3D objects, achieving state-of-the-art performance.
    \item A material diffusion model with illumination confidence to handle various lighting conditions with one model.
    \item A progressive material generation scheme guided by confidence masks, along with a UV-space material diffusion model, to generate consistent and UV-ready materials.
\end{itemize}

%% file: Sections/2_related.tex
\section{Related Work}
\noindent\textbf{3D Object Generation.} 
Previous works~\cite{poole2023dreamfusion,tang2023make,chen2024comboverse,wang2024themestation,lin2023magic3d,huang2024humannorm,sun2023dreamcraft3d,wang2023prolificdreamer} rely on image diffusion models for 3D generation with score distillation sampling but suffer from long optimization times for each object. To mitigate this, recent works have explored feed-forward models. These approaches either apply 3D diffusion models~\cite{nichol2022point,wang2023rodin,jun2023shap,hong20243dtopia} or employ U-Net or transformer-based models~\cite{hong2024lrm,li2024instant3d,he2023openlrm,tang2024lgm,zou2024triplane,wang2024phidias} to directly generate 3D representations. Despite achieving impressive results, these models are limited in their ability to generate realistic materials, often producing textures entangled with complex lighting information. This limitation hinders the adaptability of generated objects for downstream applications, where material properties are essential. Recent works, such as Clay~\cite{zhang2024clay}, Meta 3DGen~\cite{bensadoun2024meta}, and 3DTopia-XL~\cite{chen20243dtopia}, have proposed frameworks for generating 3D objects with PBR materials from prompts or images. However, these methods struggle to handle diverse 3D object inputs under different texture and illumination conditions for different applications.

\begin{figure*}[t]
    \centering
\includegraphics[width=\textwidth]{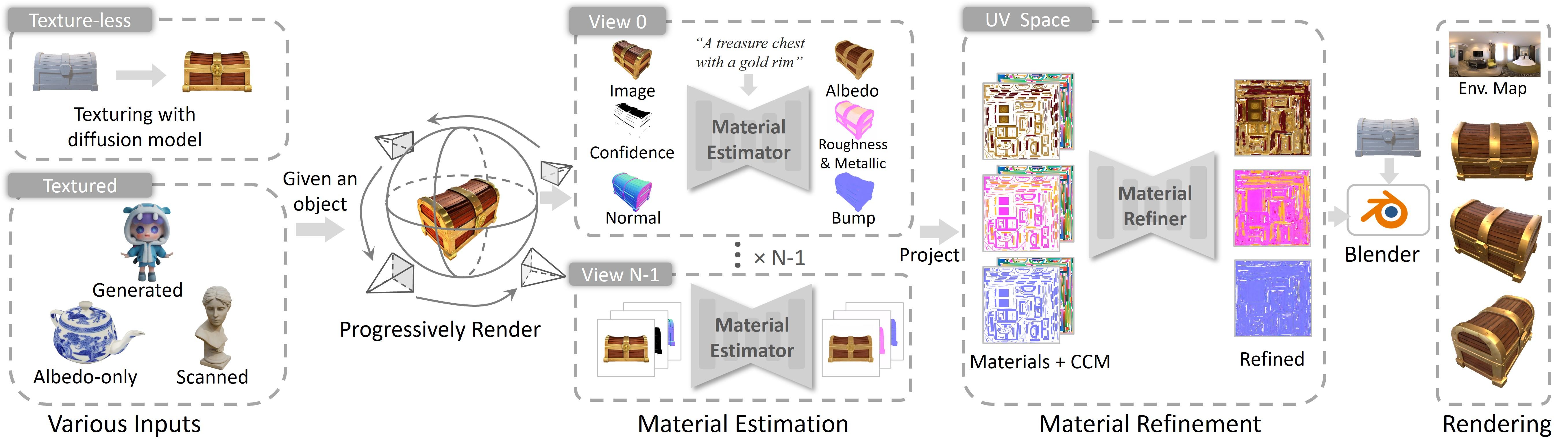}
    \caption{\textbf{Overview of Material Anything.} For texture-less objects, we first generate coarse textures using image diffusion models, similar to the texture generation method~\cite{chen2023text2tex}. For objects with pre-existing textures, we directly process them. Next, a material estimator progressively estimates materials for each view from a rendered image, normal, and confidence mask. The confidence mask serves as additional guidance for illuminance uncertainty, addressing lighting variations in the input image and enhancing consistency across generated multi-view materials. These materials are then unwrapped into UV space and refined by a material refiner. The final material maps are integrated with the mesh, enabling the object for downstream applications.}
    
    \label{fig:pipeline}
\end{figure*}

\noindent\textbf{Texture Generation for 3D Object.} Given a texture-less 3D model, TEXTure~\cite{richardson2023texture} introduces a progressive approach that generates textures view-by-view using diffusion. Latent-NeRF~\cite{metzer2023latent} further improves efficiency in a lower-dimensional latent space, enabling high-quality texture synthesis. Text2Tex~\cite{chen2023text2tex} introduces an automatic view sequence generation scheme, optimizing the generation sequence to achieve more consistent textures. SyncMVD~\cite{liu2023text} generates textures from multiple views simultaneously, ensuring coherent alignment of textures. Paint3D~\cite{zeng2024paint3d} first creates an initial texture map, then refines it in the UV space, achieving highly detailed and spatially coherent textures. Despite these advancements, the textures generated by these methods are typically entangled with complex lighting and shadows, lacking realistic material modeling.

\noindent\textbf{Material Generation for 3D Object.} 
Early works employ optimization-based methods for material generation. Fantasia3D~\cite{chen2023fantasia3d} involves learnable materials when generating 3D objects. NvDiffRec~\cite{munkberg2022extracting} jointly optimizes the topology, materials, and lighting conditions from multi-view image observations. Matlaber~\cite{xu2023matlaber} proposes a latent BRDF auto-encoder to enable material-aware 3D generation. Paint-it~\cite{youwang2024paint} introduces a re-parameterization of PBR texture maps, facilitating robust and efficient optimization. DreamMat~\cite{zhang2024dreammat} finetunes a lighting-aware diffusion model for distilling PBR materials. These methods are time-intensive and often suffer unnatural color. Retrieval-based methods~\cite{fang2024make, zhang2024mapa} rely on large multi-modal models such as SAM~\cite{kirillov2023segment} and GPT~\cite{achiam2023gpt} for segmentation and material assignment, limiting their scalability. Recently, several works~\cite{vecchio2024controlmat,vainer2024collaborative,zeng2024rgb} attempt to generate materials for images with diffusion models, however, they are not applicable to 3D objects due to consistency issues.

%% file: Sections/3_approach.tex
\section{Approach}
Material Anything, illustrated in Fig.~\ref{fig:pipeline}, is a unified framework for generating high-quality physical materials for 3D objects, accommodating various lighting and texture scenarios. It effectively handles \textbf{(1)} texture-less objects, \textbf{(2)} albedo-only objects (without lighting effects), \textbf{(3)} scanned objects (realistic lighting), and \textbf{(4)} generated objects (unrealistic lighting). Unlike existing methods that treat these scenarios as separate tasks, our method unifies them under a single framework. To this end, we reformulate 3D material generation as an image-based material estimation task, enabling the use of pre-trained image diffusion models and simplifying the overall process. Our framework centers on two core components. First, we employ a diffusion-based material estimator equipped with confidence masks, which generates materials for each view of the input object (Sec.~\ref{sec:method_estimator}). Next, we introduce a progressive material generation strategy that utilizes confidence masks to ensure consistency of generated materials across views, and further integrate a UV-space diffusion model for material refinement. (Sec.~\ref{sec:method_painting}). Finally, we provide the construction details of our Material3D dataset in Sec.~\ref{sec:dataset}.

\subsection{Image-based Material Diffusion} \label{sec:method_estimator}
The material estimator aims to produce albedo ($\mathbb{R}^3$), roughness ($\mathbb{R}$), metallic ($\mathbb{R}$), and bump ($\mathbb{R}^3$) maps from an input image. Given the limited availability of PBR data, we opt to leverage the power of pre-trained image diffusion models~\cite{rombach2022high}. However, these models are primarily designed for natural image generation, posing three challenges for PBR material generation: \textit{Channel Gap.}  Image diffusion models typically operate on three channels (RGB), while PBR materials require more than three channels (eight channels in our method). This discrepancy can lead to inaccurate material representations, as the model must adapt to producing a more complex set of outputs.  \textit{Domain Gap.} PBR material maps are different from natural images. This significant difference leads to unstable training and suboptimal performance. \textit{Various Lighting.} Finally, our material estimator must be robust across images with diverse lighting conditions, ensuring consistent performance. To address these challenges, we introduce several key components.


\begin{figure}[t]
    \centering
\includegraphics[width=\linewidth]{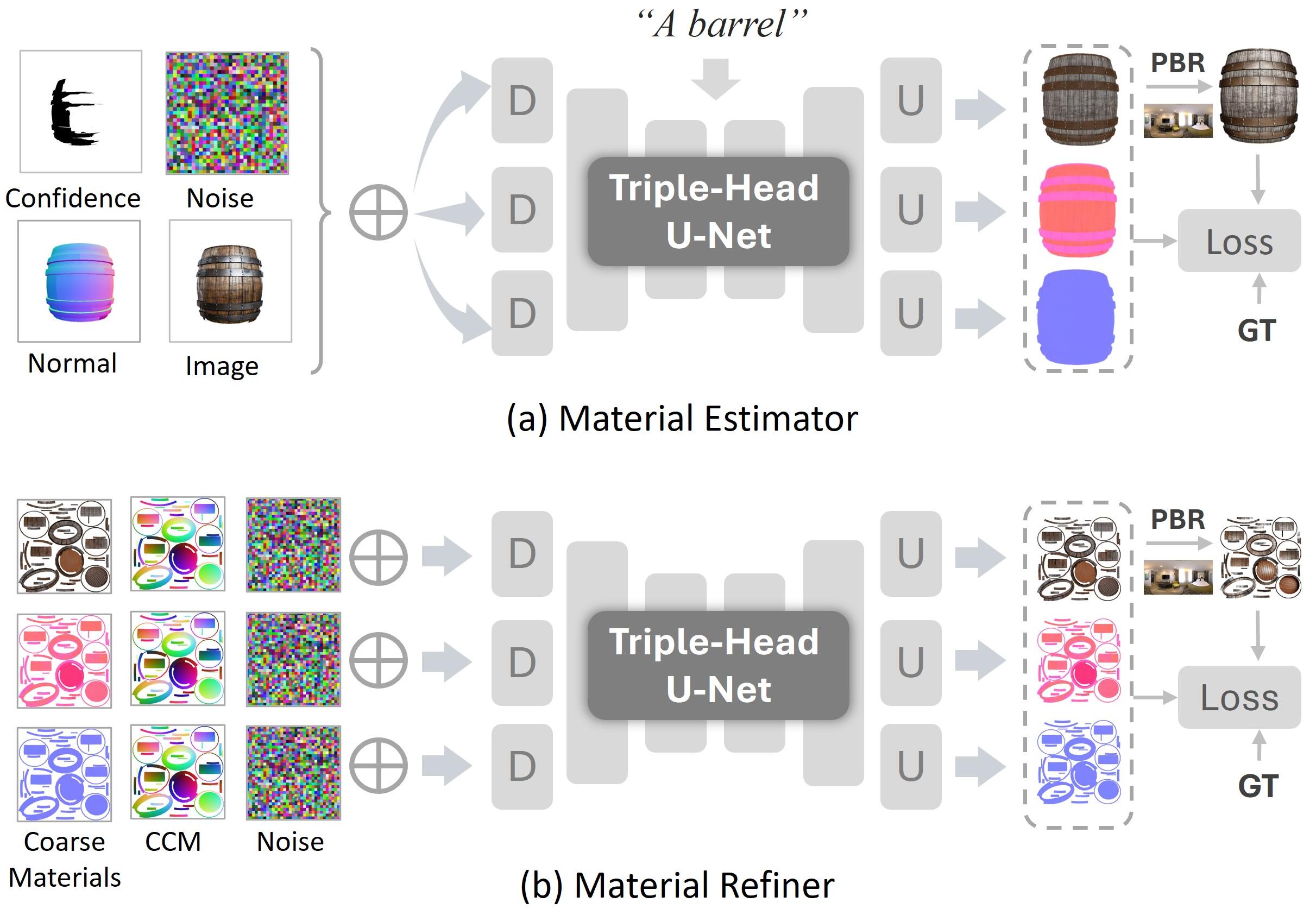}
    \caption{\textbf{Architectural design of material estimator and refiner.} Both employ a triple-head U-Net, generating albedo, roughness-metallic, and bump maps via separate branches. }
    \label{fig:network}
\end{figure}

\noindent\textbf{Triple-Head Diffusion.} To adapt the three-channel diffusion model to handle multiple material-specific channels, one solution is to train a material VAE~\cite{vainer2024collaborative}. However, this approach may discard the pre-trained priors of diffusion models, and training a customized material VAE on our limited PBR data is challenging. Inspired by previous work~\cite{zhang2024clay}, we design a triple-head U-Net architecture, illustrated in Fig.~\ref{fig:network} (a). 
The U-Net architecture comprises three distinct branches for the initial convolutional layer and first DownBlock, followed by shared middle layers enabling concurrent denoising across material modalities. The final UpBlock and output convolutional layer are also separated into three branches. Each output head produces a specific material map: an albedo map, a combined roughness-metallic map (R channel set to 1, G for roughness, B for metallic), and a bump map. This triple-head structure, instead of combining all materials into one output, ensures that each material map is generated without mutual interference while maintaining consistency among them.



\noindent\textbf{Confidence-Adaptive Generation.} To manage inputs with various lighting conditions, we categorize these conditions into two main groups: high confidence (\emph{e.g.}, scanned objects) and low confidence (\emph{e.g.}, no lighting and generated lighting). To guide the model, we introduce a certainty mask that indicates illumination confidence. For inputs with realistic lighting, the confidence value is set to 1, encouraging the diffusion model to leverage lighting cues for material estimation. In contrast, for inputs lacking lighting or with generated lighting, the confidence is set to 0, directing the model to focus on material generation instead of estimation. Note that, for images with generated lighting, the mask can selectively assign values of 1 in known material regions and 0 elsewhere to enhance multi-view material consistency, as detailed in the progressive material generation (Sec.~\ref{sec:method_painting}). The confidence mask enables the diffusion model to seamlessly transition between material estimation and generation, effectively managing both realistic and synthetic lighting scenarios. The learning objective is v-prediction~\cite{salimans2022vprediction}:
\begin{equation}
    \mathcal{L}_{v} = \mathbb{E}_{\mathbf{z}, \mathbf{c}, y, \mathbf{v}, t}
    \left\| \hat{\mathbf{V}}_{\theta} (\mathbf{z}_t; \mathbf{c}, y) - \mathbf{v}_t \right\|_2^2 ,
\label{eq:latent_loss}
\end{equation}
where $\mathbf{v}_t$ represents the v-prediction targets at time-step $t$ for the albedo, roughness-metallic, and bump maps, respectively. $\mathbf{z}_t$ denotes the noise latent. $\mathbf{c}$ denotes the conditioning inputs (input image, confidence mask, and normal map), while $y$ represents the text prompt. $\hat{\mathbf{V}}_{\theta}$ refers to our triple-head diffusion network with learnable parameters $\theta$.

\begin{figure}
    \centering
    \includegraphics[width=\linewidth]{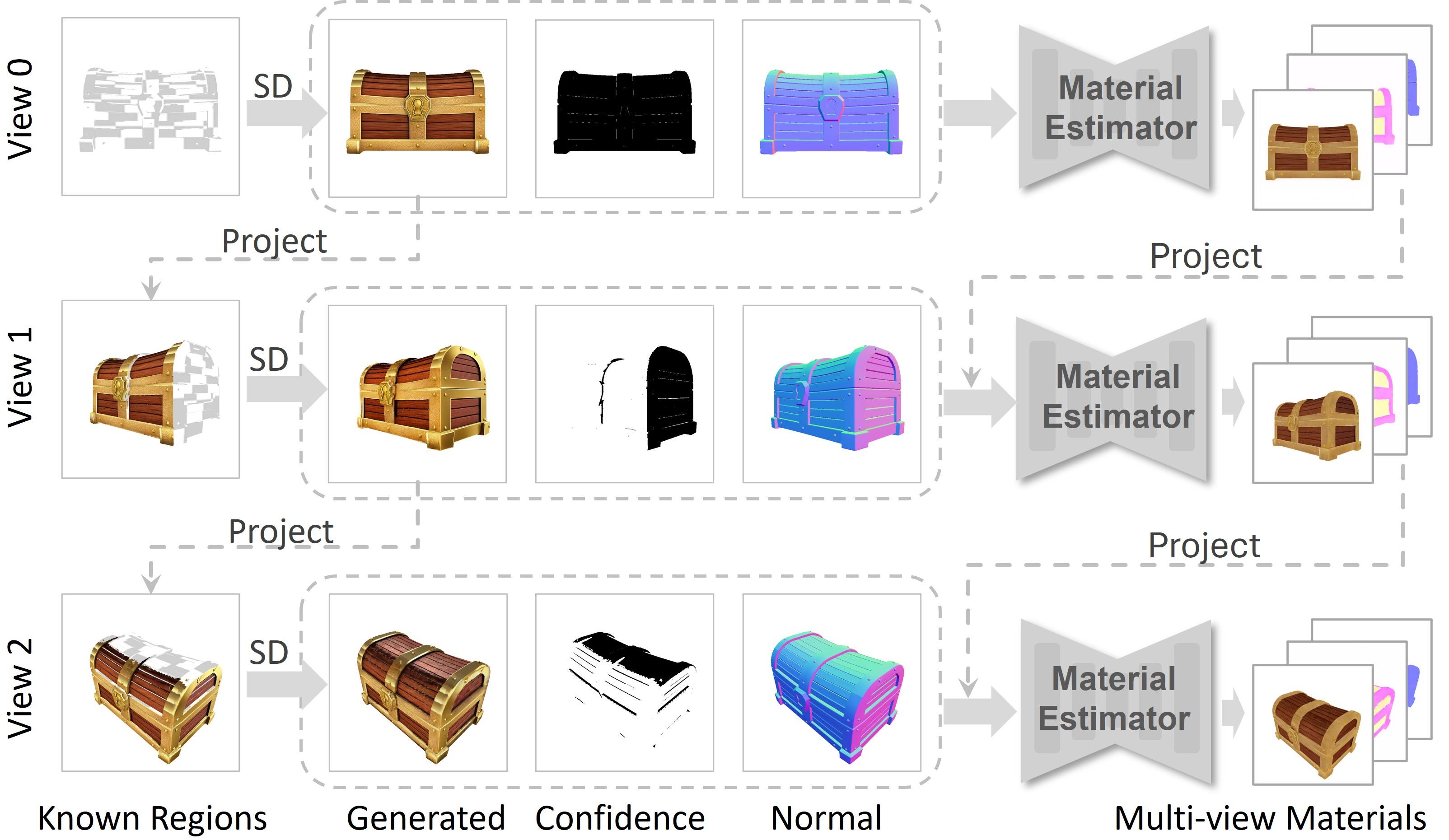}
    \caption{\textbf{Progressive material generation process for a texture-less object.} ``Project" denotes projecting known regions for the latent initialization of the next view. ``SD" denotes the pre-trained stable diffusion model~\cite{rombach2022high} with depth ControlNet~\cite{zhang2023adding} }
    \label{fig:progressive}
\end{figure}

\noindent\textbf{Rendering Loss.} Due to the significant domain gap, training the material diffusion model with only the v-prediction objective is unstable. To address this, we introduce a rendering loss by decoding albedo, roughness-metallic, and bump map from latent space into the image for reconstruction. These components are rendered into an image $\hat{r}$ under random lighting conditions using differentiable rendering~\cite{Laine2020diffrast}. We then compute the perceptual loss~\cite{johnson2016perceptual} against the ground truth rendering $r$ as follows:
\begin{equation}
    \mathcal{L}_{p} = \sum_{l} \left\| \phi_l(\hat{r}) - \phi_l({r}) \right\|_2^2 ,
\label{eq:rendering_loss}
\end{equation}
where $\phi_l$ represents VGG network~\cite{simonyan2014very}. 
The rendering loss ensures that the estimated materials exhibit realistic behavior under diverse lighting conditions, improving the quality of generated materials. Additionally, an L2 loss is applied to each material component to further improve performance.

\subsection{Materials Generation for 3D Object}
\label{sec:method_painting}
While we have successfully estimated materials for images, applying the material estimator to multiple views of a 3D object would lead to noticeable inconsistencies. One solution involves training a material estimator to simultaneously predict materials across multiple views, similar to multi-view diffusion~\cite{shi2023zero123++}. However, the increased number of views and channels poses a challenge for generating high-resolution materials. To adapt our 2D materials estimator for 3D objects, we propose a progressive generation strategy that dynamically estimates materials across different viewpoints based on the aforementioned confidence mask. We further project the multi-view materials into UV space and apply a refinement diffusion model, which completes occluded regions and refines materials, ensuring seamless and consistent materials maps.

\noindent\textbf{Progressive Material Generation with Confidence Guidance.} Given a 3D object $O$, we define a set of camera views and progressively generate materials for each view. For meshes lacking textures, we first generate a textured image $v_i$ using an image diffusion model conditioned on a depth map, similar to Text2Tex~\cite{chen2023text2tex}. For meshes with existing textures, we directly render the textured image $v_i$ from the input object. Figure~\ref{fig:progressive} illustrates the progressive material generation for a texture-less object. The next step is to estimate the material from the image $v_i$. 

As we generate materials for each view independently, our goal is to maintain consistency across views. When generating a new view, we aim for the materials to remain consistent with existing regions in previous views, rather than relying solely on the current view. To achieve this, First, we initialize the noise latent using materials from previously processed views $\{v_j | j < i\}$, with mask $\hat{m}$ indicating known regions, ensuring these regions are preserved and consistent. The latent initialization is formulated as:
\begin{equation}
    \hat{\mathbf{z}}_{t} = \hat{\mathbf{z}}_{t} \cdot (1-\hat{m}) + \mathbf{z}_t \cdot \hat{m},
\label{eq:init_latent}
\end{equation}
where $\hat{\mathbf{z}}_{t}$ represents the noise latent at time step $t$ for material maps, $\mathbf{z}_{t}$ denotes the latent of known regions.

Consistency is especially challenging in views with generated lighting due to exaggerated highlights and shadows. Therefore, for these views with generated lighting, we additionally utilize the confidence mask $m$ introduced in Sec.~\ref{sec:method_estimator} to further enhance consistency between newly generated and known regions. Specifically, we dynamically adjust the mask $m$, setting it to 1 for known regions with higher confidence and to 0 for regions requiring new generation. This approach guides the estimator to produce materials that align seamlessly with known regions, as our training data is designed to simulate these unrealistic lighting situations.

Next, we bake these material maps into the UV space according to the UV unwrapping of object $O$. After processing all generated views and materials, we obtain the coarse UV material maps $M^{uv}$ for the object.

\noindent\textbf{UV Refinement Diffusion.} Although coarse UV material maps are generated, issues such as seams (resulting from baking across different views) and texture holes (due to self-occlusion during rendering) remain. We thus refine material maps directly in UV space using a diffusion model. Unlike Paint3D~\cite{zeng2024paint3d}, which fine-tunes a diffusion model solely on albedo maps, our task is more complex, as it involves refining albedo, roughness-metallic, and bump maps. We trained a material refiner that takes the coarse material maps $M^{uv}$ as input, completing holes and smoothing seams. Additionally, a canonical coordinate map (CCM) is introduced to incorporate 3D adjacency information during the diffusion process, guiding the regions that require inpainting, as shown in Fig.~\ref{fig:network}~(b). By integrating these components, the refiner produces high-quality, consistent UV material maps.


\subsection{Material3D Dataset} \label{sec:dataset}
To train Material Anything, we build a dataset \emph{Material3D} that consists of 80K high-quality 3D objects curated from Objaverse~\cite{deitke2023objaverse}. Details of dataset construction are provided in the supplementary. For each model, we rendered multi-view material images (albedo, roughness, metallic, bump), and normal maps from 10 fixed camera positions. Additionally, UV material maps and the CCM were rendered to facilitate the training of the material refiner. To enable the model to handle diverse lighting scenarios, we incorporated various lighting conditions, including \textit{Point Lighting}, \textit{Area Lighting}, \textit{Environment Lighting}, and \textit{Without Lighting}, for rendering input images. Additionally, we designed a strategy to simulate the imperfect and inconsistent lighting conditions common during inference.

\noindent\textbf{Simulating Inconsistent Lighting.} We randomly select two images under different lighting conditions for a camera view and stitch portions of each into a composite during training. This enables a single image to exhibit two distinct lighting types, simulating the inconsistency in multi-view materials. Furthermore, we introduce degradations to one of the images, applying effects such as blurring and color shifts. A confidence mask is used to delineate the regions that have undergone degradation. The final input to the material estimator comprises the stitched image, the confidence mask, and the normal map. To train the material refiner, we randomly mask regions of the UV material maps and use these masked material maps as input. The CCM, derived from the UV mapping of 3D point coordinates, is also included. These maps guide the areas requiring inpainting and facilitate the integration of 3D adjacency information during the diffusion process. Refer to \emph{supplementary material} for more details on our dataset.

%% file: Sections/4_exp.tex
\begin{figure*}[!th]
    \centering
    \includegraphics[width=0.98\textwidth]{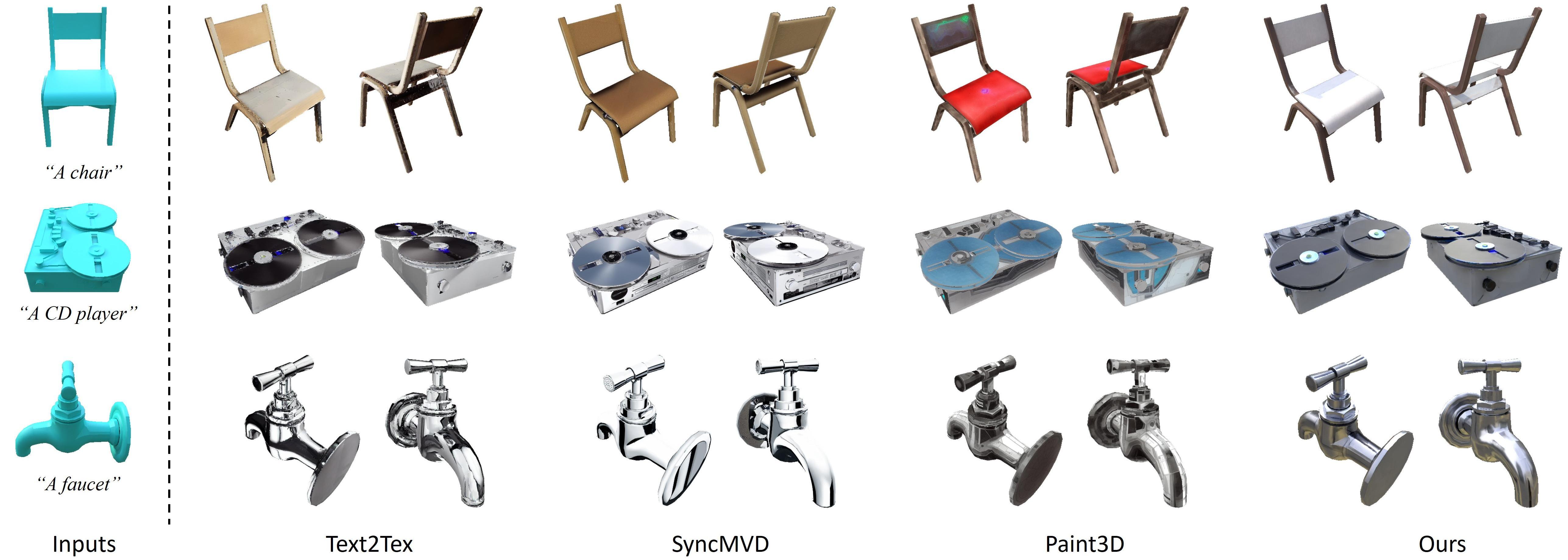}
    \caption{\textbf{Comparisons with texture generation methods.} These methods directly paint texture-less objects using image diffusion models but fail to generate the corresponding material properties.}
    \label{fig:comp_learning}
\end{figure*}

\begin{figure*}[!th]
    \centering
    \includegraphics[width=0.97\textwidth]{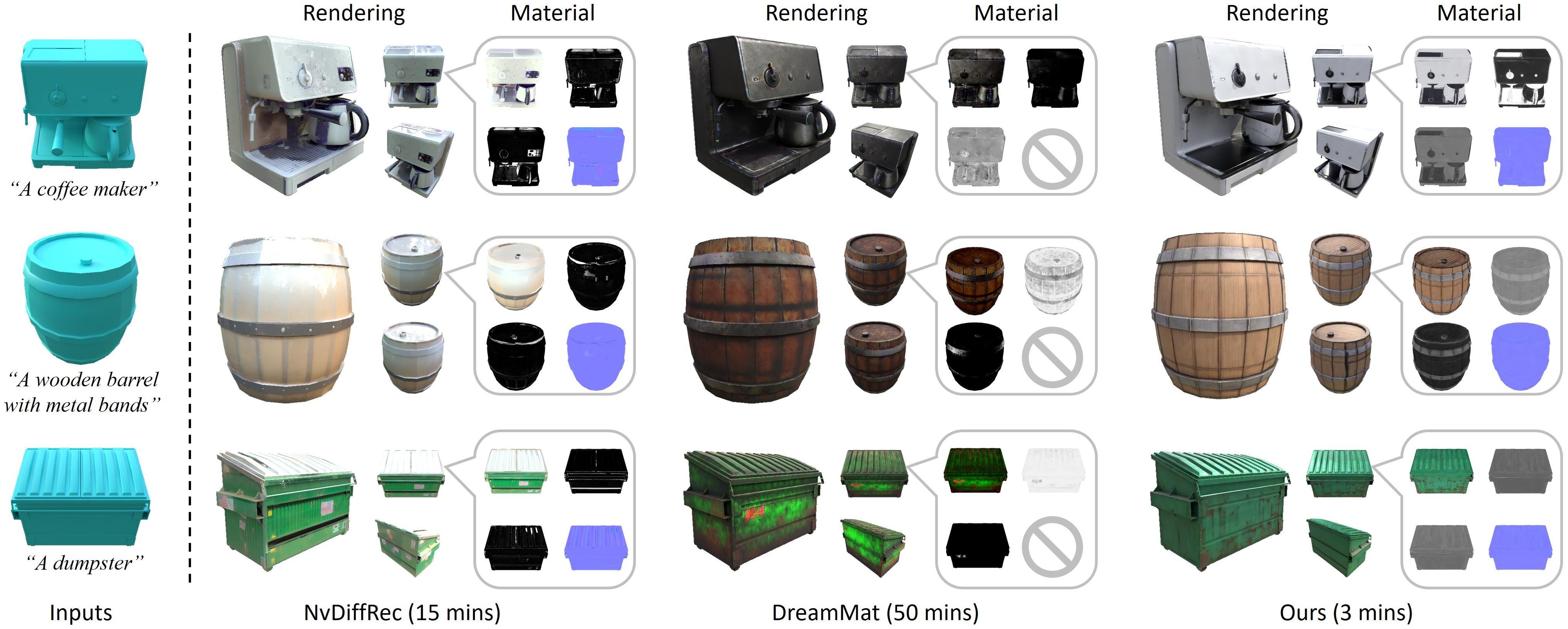}
    \caption{\textbf{Comparisons with optimization methods.} NvDiffRec~\cite{munkberg2022extracting} estimates materials using the textured model by SyncMVD~\cite{liu2023text} as input. The materials include albedo (top left); roughness (top right); metallic (bottom left); bump (bottom right).  }
    \label{fig:comp_opt}
\end{figure*}

\section{Experiments}
We compare our method with texture generation methods, Text2Tex~\cite{chen2023text2tex}, SyncMVD~\cite{liu2023text}, and Paint3D~\cite{zeng2024paint3d}. Additionally, we assess our method alongside optimization-based material generation approaches, NvDiffRec~\cite{munkberg2022extracting} and DreamMat~\cite{zhang2024dreammat}, and a retrieval-based method, Make-it-Real~\cite{fang2024make}. Finally, we also include comparisons with the closed-source methods, Rodin Gen-1~\cite{rodin} and Tripo3D~\cite{tripo3d}.

\subsection{Qualitative Evaluation}

\begin{figure}
    \centering
    \includegraphics[width=\linewidth]{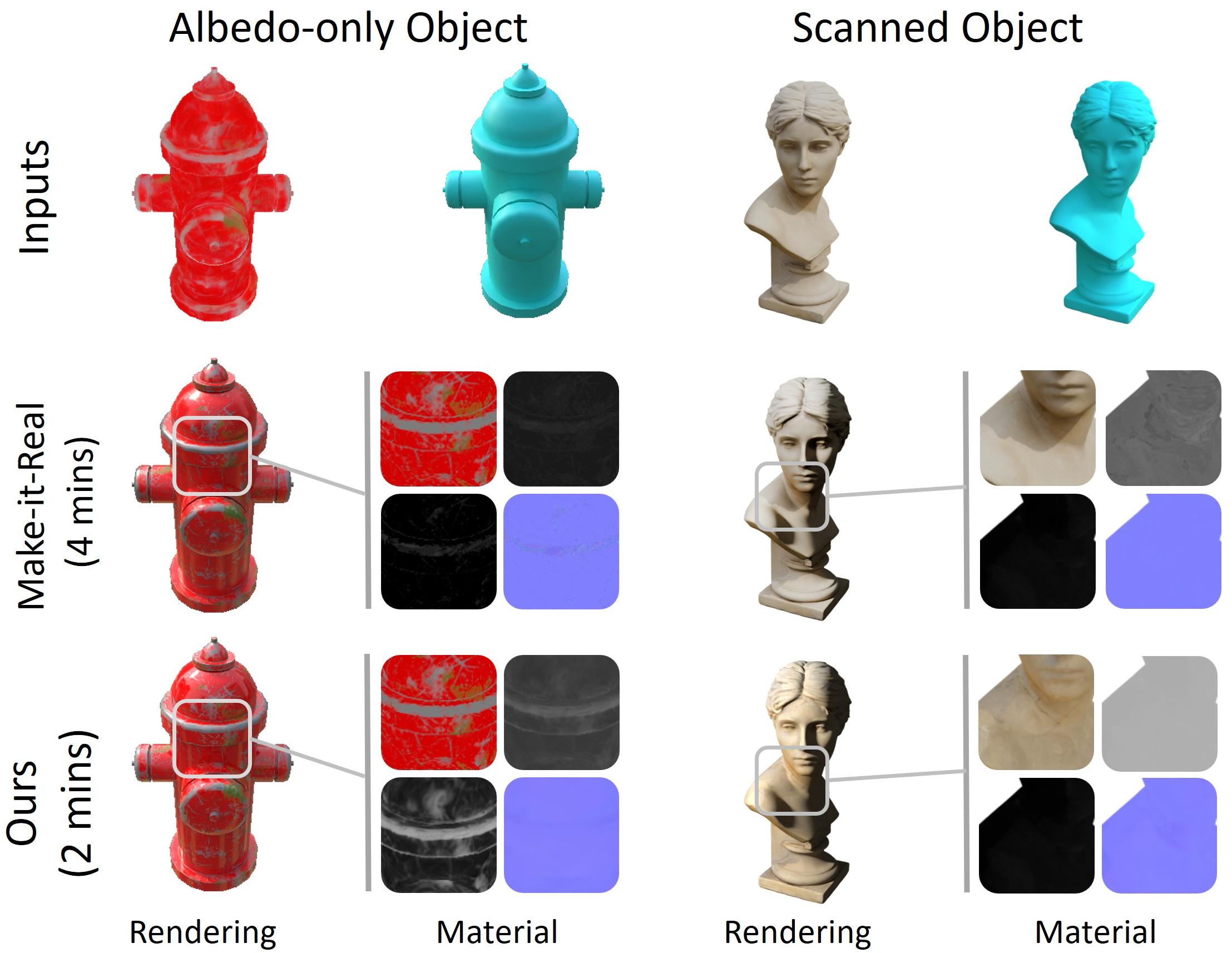}
    \vspace{-5pt}
    \caption{\textbf{Comparisons with retrieval methods.} The inputs are textured objects, including an albedo-only object and a scanned object. The materials include albedo (top left); roughness (top right); metallic (bottom left); bump (bottom right). }
    \label{fig:comp_retrieve}
\end{figure}

\begin{figure}
    \centering
\includegraphics[width=\linewidth]{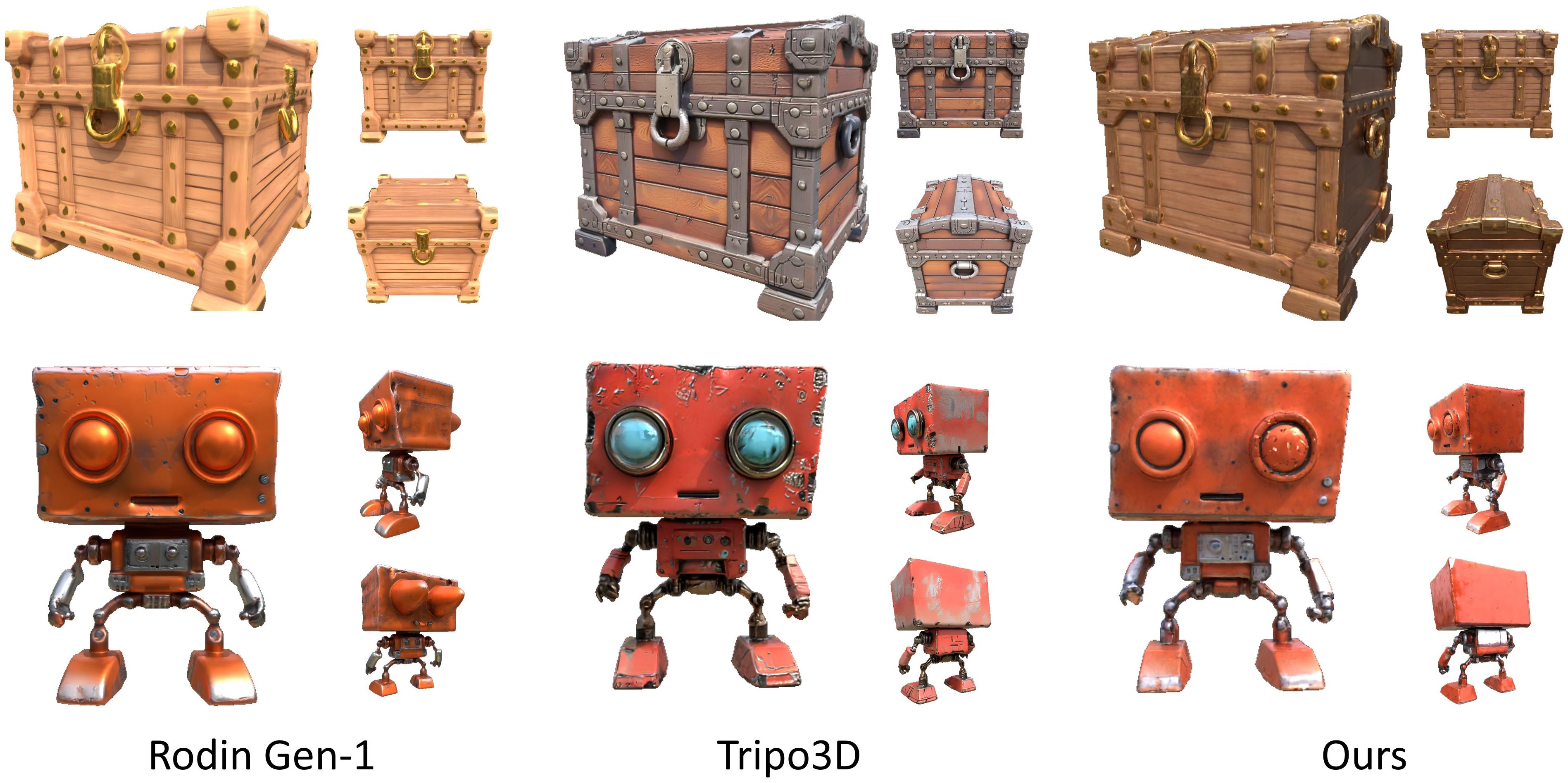}
    \vspace{-10pt}
    \caption{\textbf{Comparisons with Rodin Gen-1 and Tripo3D.} Rodin Gen-1 and Tripo3D are two closed-source methods. Our approach uses significantly less data, yet produces comparable results. }
    \label{fig:comp_others}
\end{figure}

\noindent\textbf{Comparisons with Texture Generation Methods.} We compare Material Anything with various texture generation methods in Fig.~\ref{fig:comp_learning}. These methods employ a similar strategy,  painting the texture-less meshes with a pre-trained image diffusion model. However, the lighting information generated by the diffusion model often results in textures that exhibit significant artifacts, as they are influenced by complex lighting conditions from multiple generated images. 
In contrast, our method produces clearer textures that inherently incorporate material properties, thus enabling robust support for downstream applications. 

\noindent\textbf{Comparisons with Optimization Methods.} We compare our approach with optimization methods, as shown in Fig.~\ref{fig:comp_opt}. These methods require extensive optimization time for each object and have difficulty accurately identifying object materials. In contrast, our method effectively distinguishes materials, as demonstrated in the barrel example, where it accurately represents the metal bands and wooden planks. This capability underscores the superiority of our method in generating realistic and diverse material maps.

\noindent\textbf{Comparisons with Retrieval Methods.} For input objects with existing textures, we compare ours with the retrieval method Make-it-Real, as shown in Fig.~\ref{fig:comp_retrieve}. Make-it-Real retrieves materials based on segmenting the initial texture, which presents several limitations. First, the segmentation process struggles with accurately capturing small regions, such as the peeling paint on the fire hydrant. Additionally, it encounters difficulty in removing shadows in the initial texture, as observed in the shadowed albedo of the sculpture example. In contrast, our method generates more accurate material, better preserving fine details and removing artifacts such as shadows.

\noindent\textbf{Comparisons with Tripo3D and Rodin Gen-1.} We compare our method with two closed-source methods, Tripo3D and Rodin Gen-1, as shown in Fig.~\ref{fig:comp_others}. We utilize texture-less meshes generated by Tripo3D as input for our material generation. Additionally, we provide the front-view image by an image diffusion model to Rodin Gen-1, ensuring the generation of the same 3D objects. While both Tripo3D and Rodin Gen-1 utilize significantly larger-scale training datasets, our method produces comparable results.

\subsection{Quantitative Evaluation}
The quantitative evaluation of our method is presented in Tab.~\ref{tb:quantitative_evaluation}. As shown, our method achieves a lower FID score, indicating that our generated textures are closer in distribution to those in Objaverse. Furthermore, the higher CLIP score of our method demonstrates its capability to generate textures more accurately aligned with the prompts.


\begin{table}[!tb]
\scriptsize
\centering
\caption{\textbf{Quantitative comparisons.} FID and CLIP scores (similarity between rendered views and text prompts) are computed on 1,200 images from 20 textured objects. For comparison with Make-it-Real, the CLIP score is calculated between rendered images from generated textures and those in Objaverse.}
\begin{tabularx}{\linewidth}{@{}lccCc}
\toprule
 Method &Type &Input Mesh & FID $\downarrow$ &CLIP Score $\uparrow$ \\
 \midrule
Text2Tex~\cite{chen2023text2tex} &Learning & Texture-less &116.41  & 30.33    \\
SyncMVD~\cite{liu2023text}  &Learning & Texture-less  &118.46  &  30.66   \\
Paint3D~\cite{zeng2024paint3d} &Learning & Texture-less  &153.20  &  28.40   \\
NvDiffRec~\cite{munkberg2022extracting}  &Optimization & Texture-less &103.81  &  30.14   \\
DreamMat~\cite{zhang2024dreammat} &Optimization & Texture-less  &113.34  &  30.64   \\
Ours &Learning & Texture-less  &\textbf{100.63}  & \textbf{31.06}    \\
 \midrule
Make-it-Real~\cite{fang2024make} &Retrieval &  Textured & 104.38  & 88.62    \\
Ours &Learning & Textured &\textbf{101.19}  & \textbf{89.70}    \\
\bottomrule
\end{tabularx}
\label{tb:quantitative_evaluation}
\end{table}

\subsection{Ablation Study}

\begin{table}[!tb]
\scriptsize
\centering
\caption{\textbf{Ablation study for triple-head U-Net and rendering loss}.  RMSE is calculated for the materials across the views from 1,000 Objaverse objects.}
\begin{tabularx}{\linewidth}{@{}lCcC@{}}
\toprule
 Materials & W/O Triple-head & W/O Rendering Loss & Full \\
\midrule
Albedo  &0.0800  &0.1442   & \textbf{0.0604}   \\
Roughness & 0.1196   & 0.1943  &  \textbf{0.0877}  \\
Metallic &0.1584   &0.2594   & \textbf{0.1193}  \\
Bump &0.0824   &0.0716   & \textbf{0.0313}   \\
\bottomrule
\end{tabularx}
\label{tb:ablation}
\end{table}


\begin{table}[!tb]
\scriptsize
\centering
\caption{\textbf{Ablation study for confidence masks}. Mean RMSE is calculated for materials from 1,000 Objaverse objects with different simulated lighting conditions, including light-less (albedo-only), realistic (scanned), and unrealistic light (generated).}
\begin{tabularx}{\linewidth}{@{}lCCCC@{}}
\toprule
  & Light-less & Realistic  & Unrealistic & Mean \\
\midrule
W/O Confidence  & 0.1521  & 0.1074 & 0.1111 & 0.1235 \\
Full            & \textbf{0.1102}  & \textbf{0.0747} & \textbf{0.0847} & \textbf{0.0899}\\
\bottomrule
\end{tabularx}
\label{tb:ablation_mask}
\end{table}

\begin{figure}
    \centering
    \includegraphics[width=\linewidth]{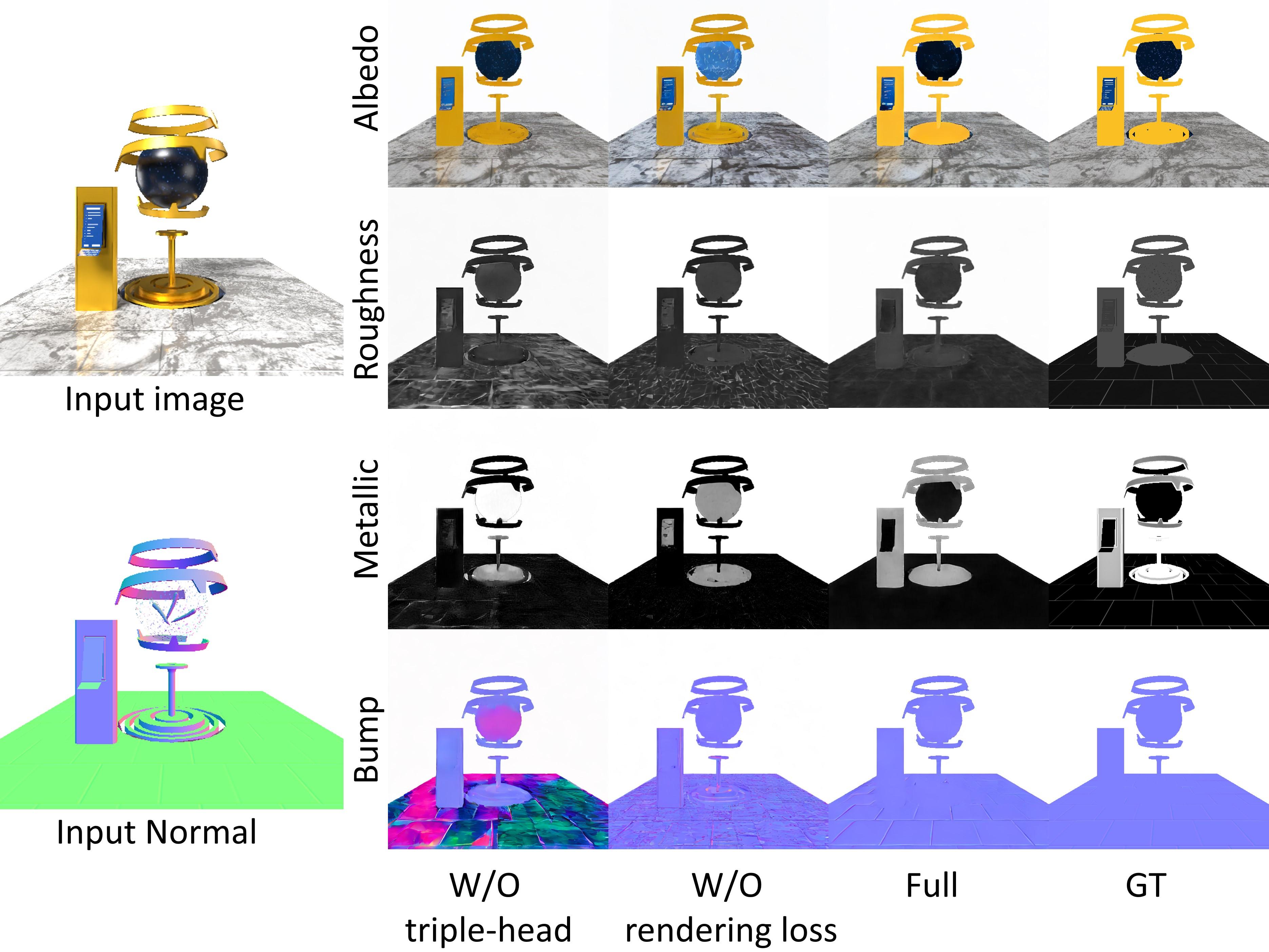}
    \caption{\textbf{Effectiveness of triple-head U-Net and rendering loss.} In both ablation experiments, the confidence mask is set to 1.}
    \label{fig:ablations_view}
\end{figure}

\textbf{Effectiveness of Triple-Head U-Net.}  We evaluate the performance of our method using a vanilla U-Net architecture that directly generates all materials as a 12-channel latent, instead of a triple-head U-Net. As shown in Tab.\ref{tb:ablation}, the performance degrades due to the coupling effect between materials when outputting a single 12-channel latent. In Fig.~\ref{fig:ablations_view}, this coupling effect is evident, where the bumps are incorrectly colored due to interference from the albedo. In contrast, the triple-head U-Net effectively decouples the materials. Additionally, the shared backbone among the three heads ensures alignment across the different material maps.

\noindent\textbf{Effectiveness of Rendering Loss.} In Tab.~\ref{tb:ablation}, we present the quantitative results of our method when trained without rendering loss. Notably, performance in this ablation is worse compared to the variant trained with rendering loss. As illustrated in Fig.~\ref{fig:ablations_view}, the version without rendering loss exhibits noticeable detail degradation, with visible artifacts across the material views. Rendering loss acts as an additional constraint in image space, ensuring consistency under varying lighting conditions, which enhances training stability and aids in capturing finer material properties. The results highlight the critical role of rendering loss in enhancing the fidelity and stability of our material estimator.

\begin{figure}
    \centering
    \includegraphics[width=\linewidth]{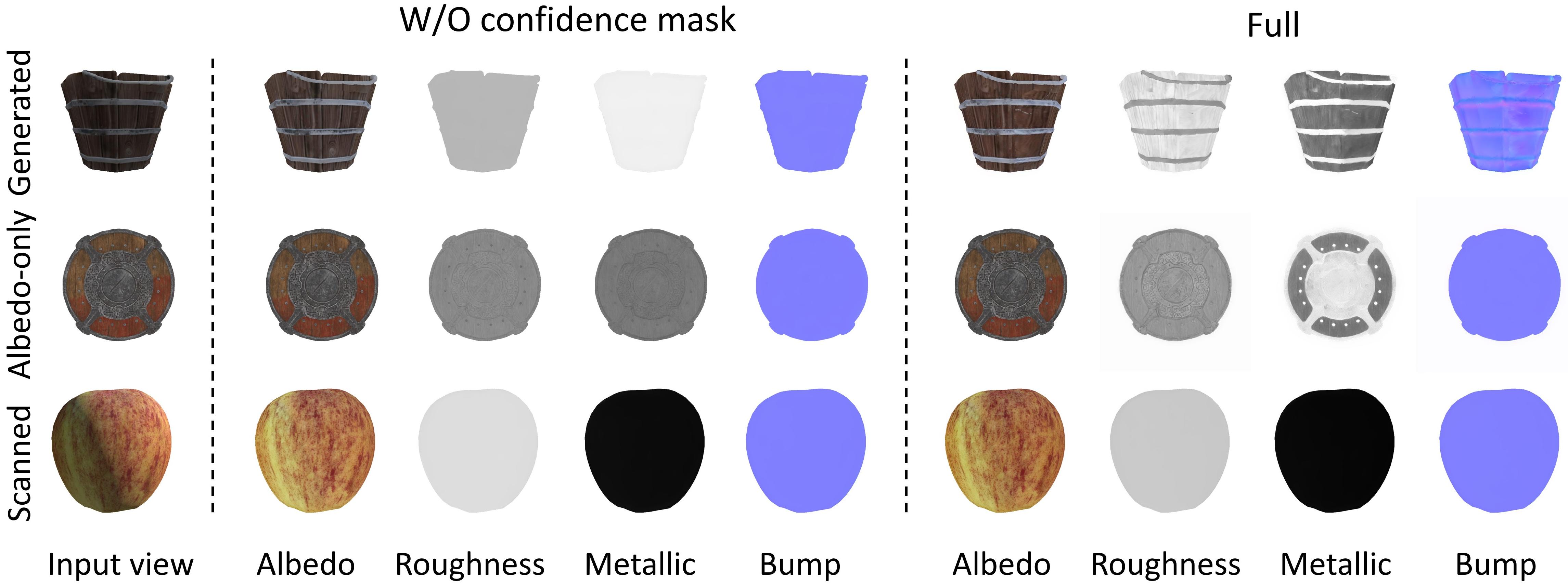}
    \caption{\textbf{Effectiveness of confidence mask for various lighting conditions.} ``W/O confidence mask'' indicates results from the material estimator without the confidence mask as input.}
    \label{fig:ablations_mask}
    \vspace{-4mm}
\end{figure}

\begin{figure}
    \centering
    \includegraphics[width=\linewidth]{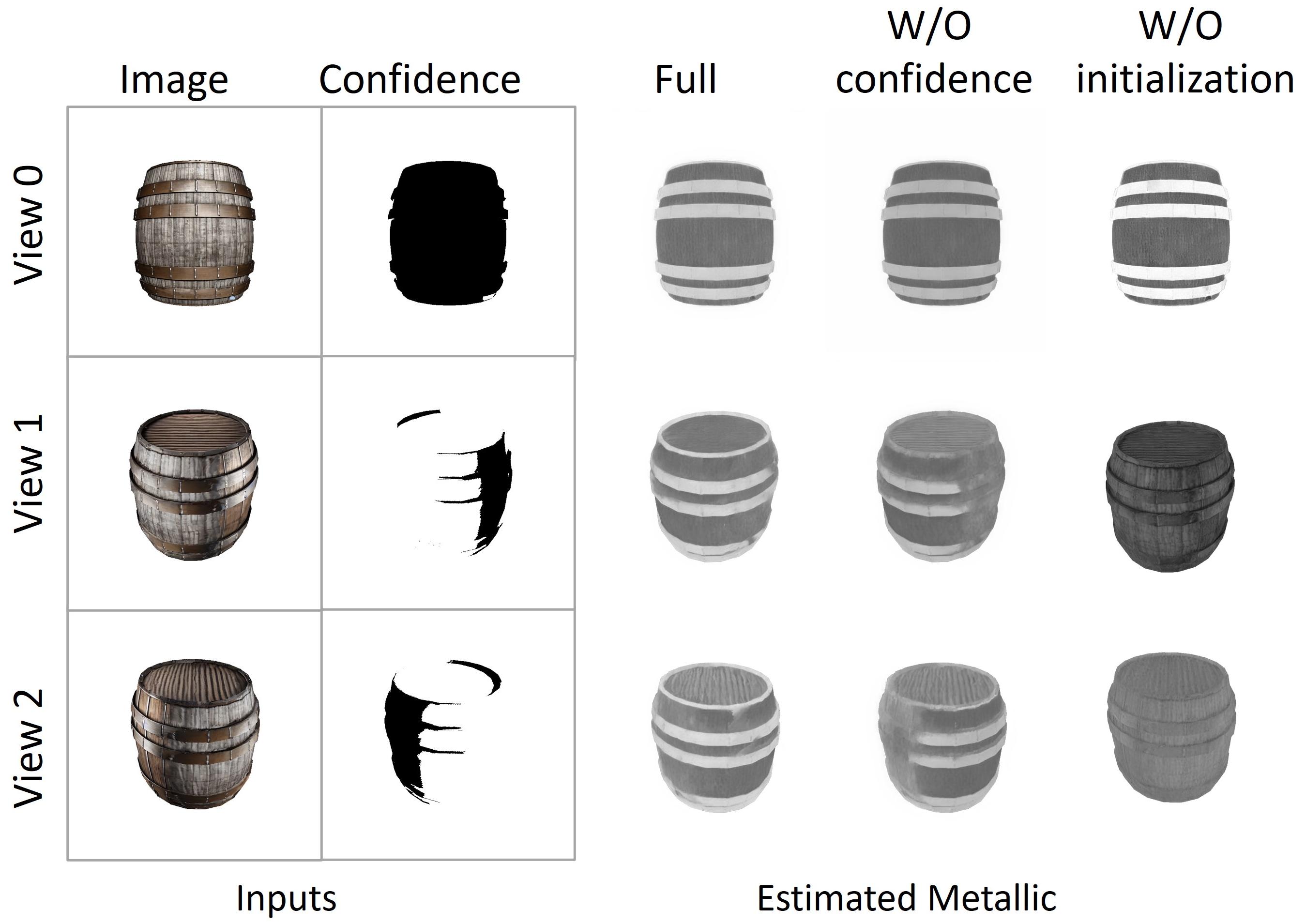}
    \caption{\textbf{Effectiveness of strategies for material consistency.} }
    \label{fig:ablations_multi_view}
\end{figure}

\noindent\textbf{Effectiveness of Confidence Mask.}
As shown in Fig.~\ref{fig:ablations_mask}, the material estimator without confidence masks struggles to generate high-quality materials under different lighting conditions. In contrast, when guided by the confidence mask, the model adapts well to these input variations. Table~\ref{tb:ablation_mask} presents quantitative results for the model without confidence masks on Objaverse objects across different lighting conditions, revealing significant drops in material accuracy.
Furthermore, for objects with generated lighting, the progressive generation without the confidence mask also shows noticeable inconsistencies in the material maps, as shown in Fig.~\ref{fig:ablations_multi_view}. Conversely, with the confidence mask employed, the model can distinguish between regions for estimation and those for generation. By guiding the training process to focus on relevant regions, our method achieves more consistent, artifact-free materials. These results demonstrate that the confidence mask improves material consistency and addresses variations in lighting.

\begin{figure}
    \centering
    \includegraphics[width=\linewidth]{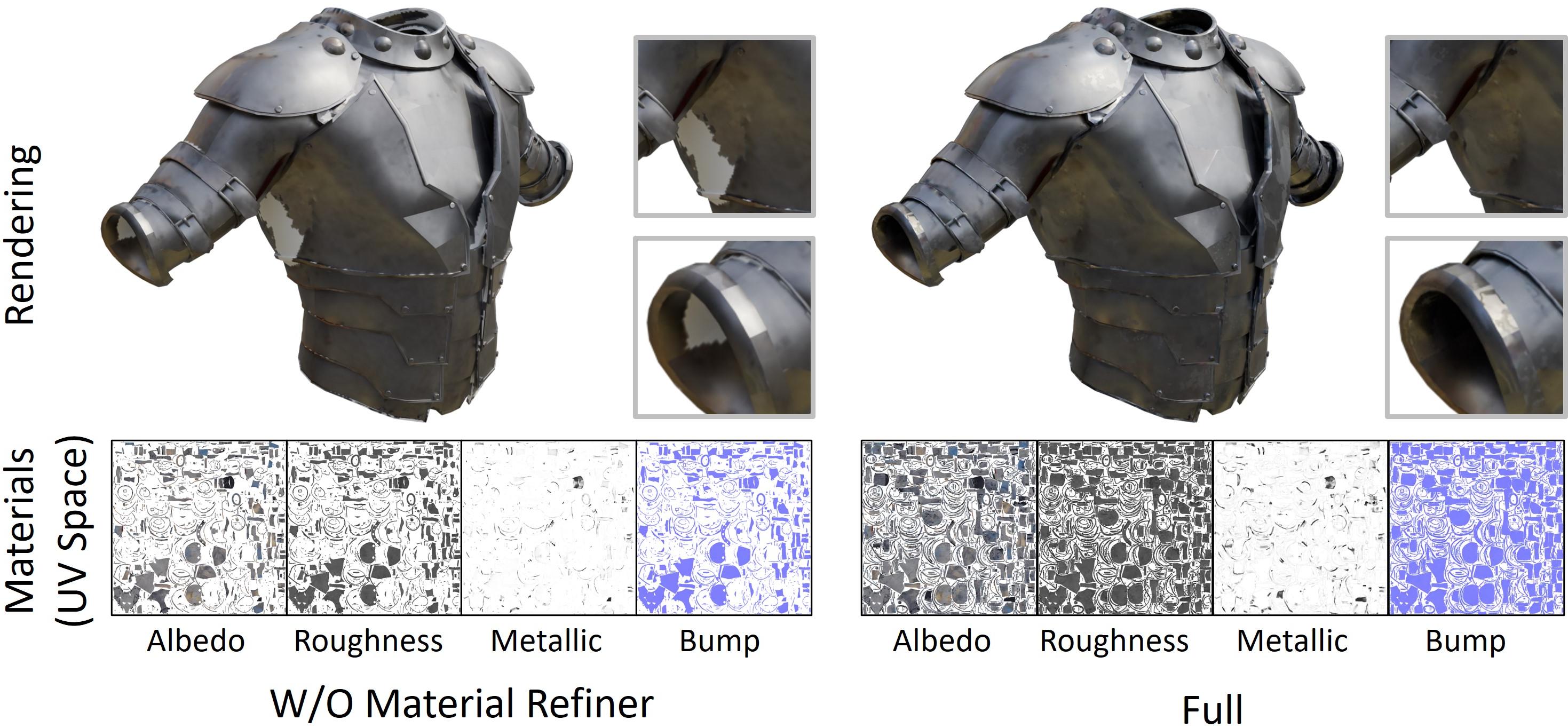}
    \caption{\textbf{Effectiveness of the UV-space material refiner.} The material refiner effectively fills in holes caused by occlusions.}
    \label{fig:ablations_uv}
\end{figure}

\noindent\textbf{Effectiveness of Known Material Initialization.} Figure~\ref{fig:ablations_multi_view} shows the results of our method without using known materials from other views for initialization. As shown, the predicted metallic properties display noticeable variations across different views. In contrast, by progressively generating materials based on known ones, our method produces more consistent materials across multiple views.




\noindent\textbf{Effectiveness of the UV-Space Material Refiner.} In Fig.~\ref{fig:ablations_uv}, we shown the results without UV refinement. As shown, several holes appear in the predicted materials due to self-occlusions, leading to incomplete material maps. After applying our material refiner, these holes are effectively filled, resulting in a more seamless and complete material representation. Our material refiner can handle occlusions and enhance the overall material generation quality.

%% file: Sections/5_app.tex

%% file: Sections/6_conclusion.tex
\section{Conclusion}
We proposed Material Anything, a unified framework to generate PBR materials for various 3D objects, including texture-less, albedo-only, generated, and scanned meshes. By leveraging a well-designed material diffusion model, our approach can generate high-fidelity materials in a feed-forward manner. To unify various input objects under complex lighting conditions, we introduced a mask to indicate confidence levels for different illuminations, which also enhances multi-view material consistency. Extensive experiments have demonstrated that our method can generate high-quality PBR materials for various objects, with a clear improvement over existing methods.

%% file: Sections/X_suppl.tex
\clearpage
\setcounter{page}{1}
\maketitlesupplementary
\appendix

\section{Material3D Dataset}
\subsection{Data Construction}
To train Material Anything, we constructed a dataset \emph{Material3D} that consists of 80K high-quality 3D objects curated from the Objaverse dataset, focusing specifically on models with comprehensive material maps. We further filtered objects using Blender, ensuring the presence of essential material maps: base color, roughness, metallic, and bump. Only models containing all these material properties were retained. Each object was imported into Blender, where Blender’s Smart UV Project tool was used to generate UV mappings. For objects with multiple parts, all components were merged into a single mesh to ensure UV mapping could be projected onto a unified 2D map. After this filtering and preparation process, we rendered multi-view material images (albedo, roughness, metallic, and bump) from 10 fixed camera positions that are consistent with the setup used in our material generation phase, which served as training data for the material estimator. In addition, we rendered images under varying lighting conditions and included normal maps as input for model training, providing diverse lighting and surface information. UV material maps and CCM were also rendered to facilitate the training of the material refiner. 
\begin{figure}
    \centering
    \includegraphics[width=0.95\linewidth]{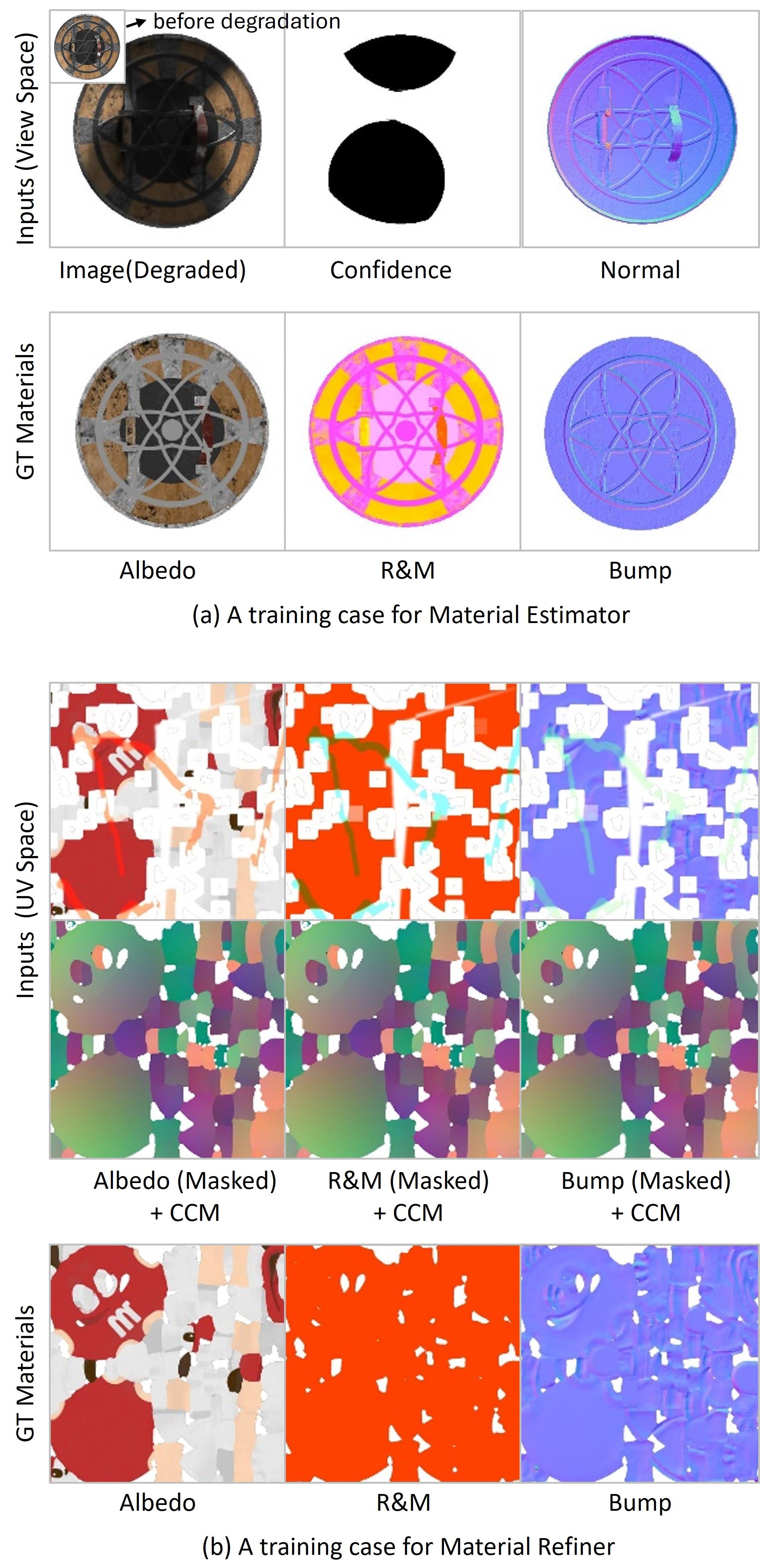}
    \caption{\textbf{The virtualization of our training data.} We apply various degradations and simulate inconsistent lighting effects in the inputs to enhance the robustness of our method.}
    \label{fig:training_case}
\end{figure}

\begin{figure}[t]
    \centering
    \includegraphics[width=\linewidth]{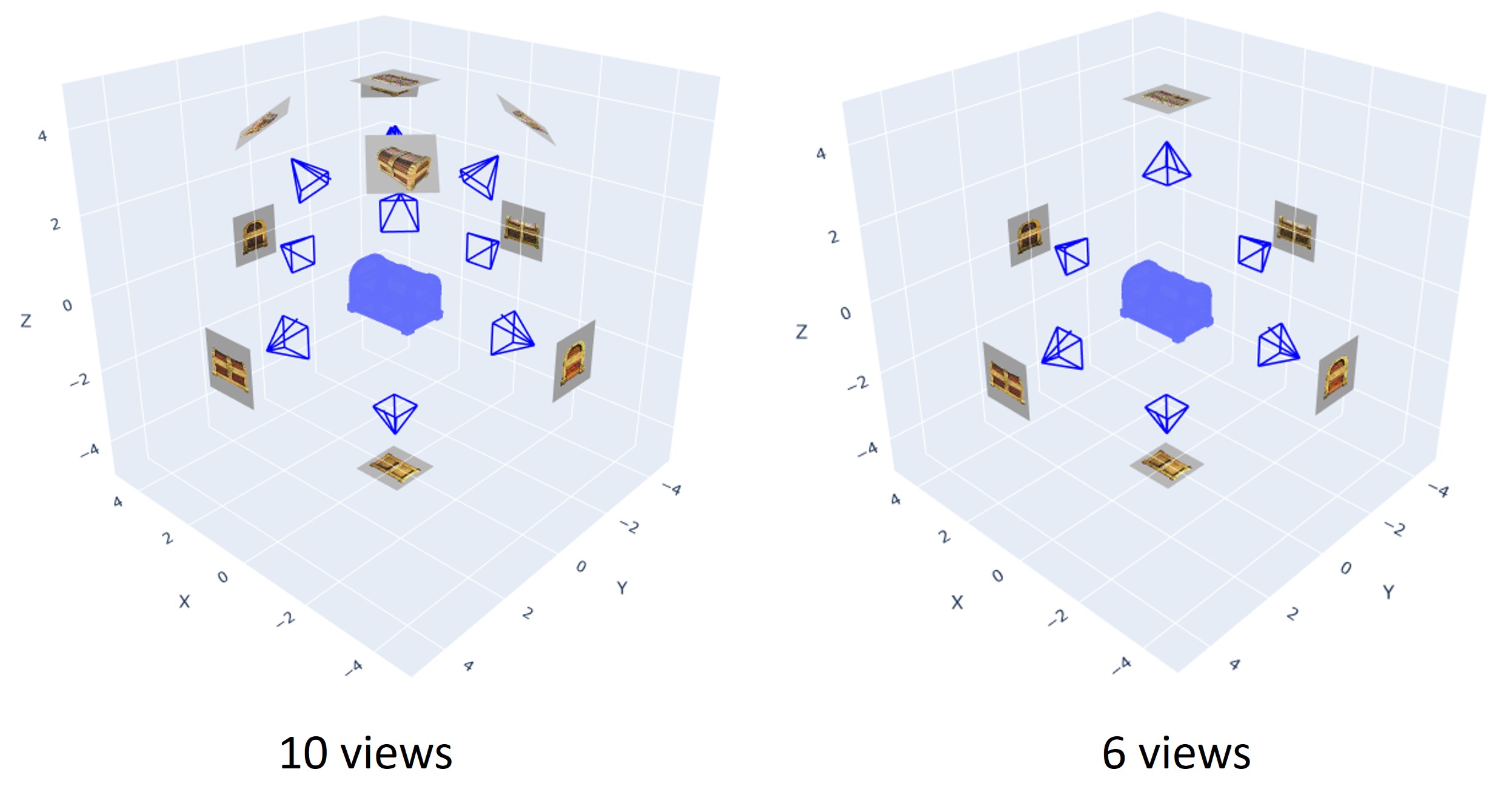}
    \caption{\textbf{The camera poses for progressive material generation and building training data.} }
    \label{fig:camera}
\end{figure}

\subsection{Lighting Conditions} Inspired by the image relighting method~\cite{zeng2024dilightnet},  we incorporated multiple lighting categories for rendering input images, enabling the model to handle diverse lighting scenarios.
\begin{enumerate} 
\item \textit{Point Lighting.} Point light sources are uniformly sampled from a hemisphere (with $0^\circ \leq \theta \leq 60^\circ$) surrounding the object, with a radius sampled in the range [4m, 5m]. The number of point lights is randomly sampled between [1, 3]. The sum power of all lights is uniformly chosen within [900W, 2400W]. To ensure the visibility of highlighted regions, the hemisphere is rotated according to the camera position, while the camera itself remained fixed at the top of the hemisphere. 
\item \textit{Area Lighting.} Similar to point lighting, area light sources are sampled from a hemisphere (with $0^\circ \leq \theta \leq 60^\circ$) with a radius from 4m to 5m. The size of the area light ranges from 3m to 10m, and its power is uniformly selected within [1000W, 2000W]. Only one area light is utilized during rendering. 
\item \textit{Environment Lighting.} Environmental lighting broadly influences scene illumination beyond isolated light sources. To counter the white balance bias common in diffusion-generated images, we employ white environment lighting with strengths ranging from [0.5, 3], avoiding colored HDR environment maps.
\item  \textit{Without lighting.} To ensure the material estimator can accurately predict materials independent of lighting conditions, we render views of objects using only albedo textures, which is the same as rendering multi-view albedo maps. 
\end{enumerate}
For each camera position, we render 13 images (including albedo, roughness, metallic, bump, normal maps, and 8 RGB images under point, area, and environment lighting). For UV material map rendering, we utilize Blender's smart UV project to unwarp the mesh, producing five UV space maps (albedo, roughness, metallic, bump, and canonical coordinate maps).

\subsection{Simulating Inconsistent Lighting Effects} To improve the robustness of the material estimator, we randomly select two images under different lighting conditions for a camera view and stitch portions of each into a composite during training. This enables a single image to exhibit two distinct lighting types, simulating the inconsistency in multi-view materials. Furthermore, we introduce degradations to one of the images, applying effects such as blurring and color shifts. A confidence mask is used to delineate the regions that have undergone degradation. The final input to the material estimator comprises the stitched image, the confidence mask, and the normal map, as shown in Fig.~\ref{fig:training_case}~(a). To train the material refiner, we randomly mask regions of the UV material maps and apply degradations such as blurring and color shifts. These masked material maps are taken as input, as shown in Fig.~\ref{fig:training_case}~(b). The CCM, derived from the UV mapping of 3D point coordinates, is also included. These maps guide the areas requiring inpainting and facilitate the integration of 3D adjacency information during the diffusion process. 

\section{Implementation Details}
\subsection{Training Details} We implemented Material Anything using the Diffusers~\cite{diffusers}, with Stable Diffusion v2.1~\cite{rombach2022high} serving as the backbone diffusion model. The training process leverages the AdamW optimizer with a learning rate of $5\times10^{-5}$. Our material estimator was trained over 300K iterations on 8 NVIDIA A100 GPUs with a batch size of 32, requiring approximately 5 days to complete. In parallel, the material refiner was trained for 150K iterations under the same GPU configuration and batch size, with a training duration of about 2 days. Training data was rendered at a resolution of $512\times512$ using Blender’s Cycles path tracer, ensuring high-quality reference materials for robust learning.
\subsection{Material Generation Details} During material generation, each input object is centered within a normalized bounding box. To capture comprehensive material properties, 6 or 10 views are rendered, as illustrated in Fig.~\ref{fig:camera}. The input image resolution for our material refiner is set to $768\times768$, while the resolution for UV material maps is $1024\times1024$. This setup ensures high-fidelity material maps that are detailed and adaptable across different viewing angles. For the input objects without UV mappings, xatlas~\cite{xatlas} is used to unwrap them. All results, including those from our method and the baselines, are generated on a single NVIDIA A100 GPU.

\begin{figure}[t]
    \centering
    \includegraphics[width=\linewidth]{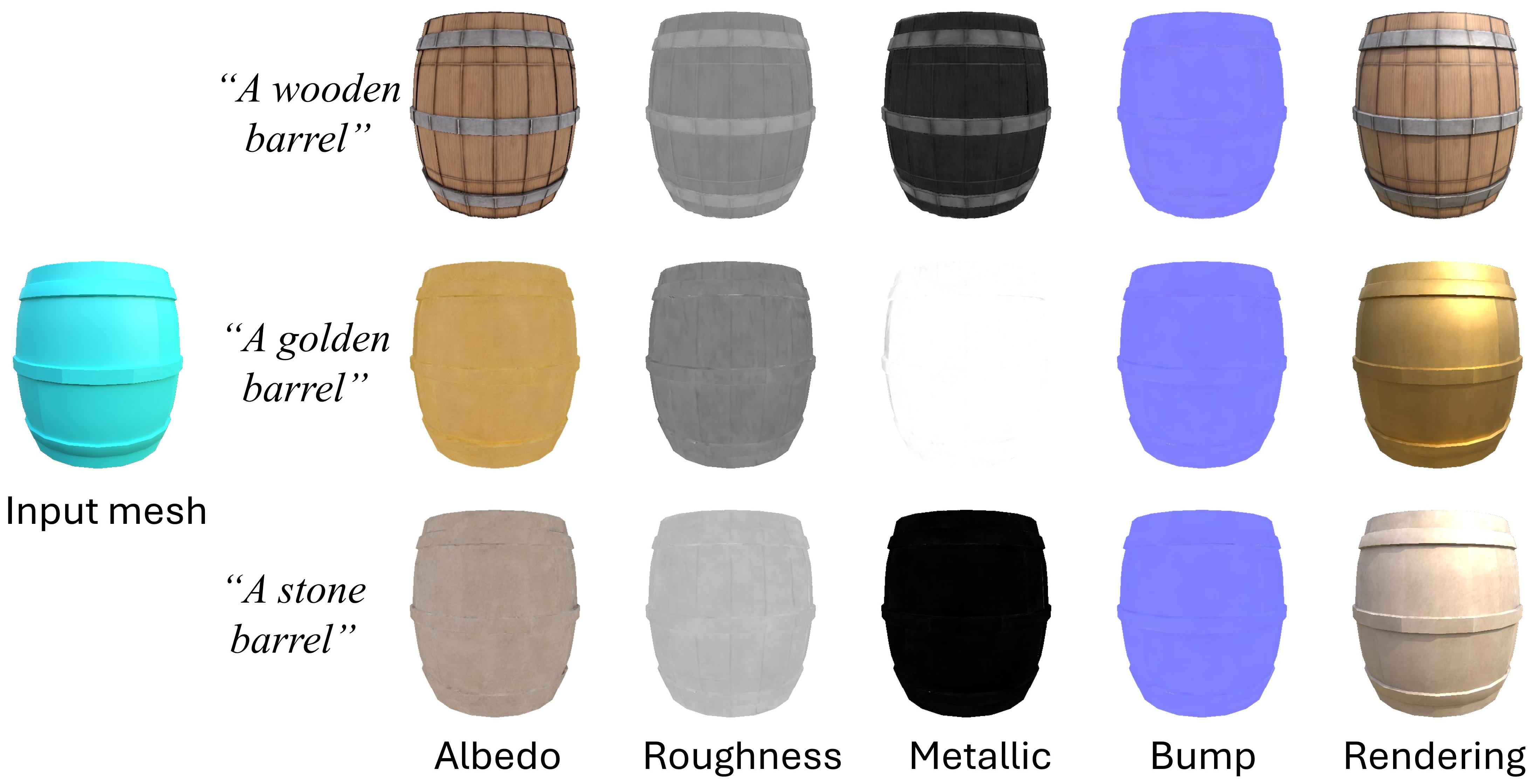}
    \caption{\textbf{Material editing with prompts.} Material Anything enables flexible editing and customization of materials for texture-less 3D objects by simply adjusting the input prompt. }
    \label{fig:editing}
\end{figure}

\begin{figure*}[t]
    \centering
    \includegraphics[width=\textwidth]{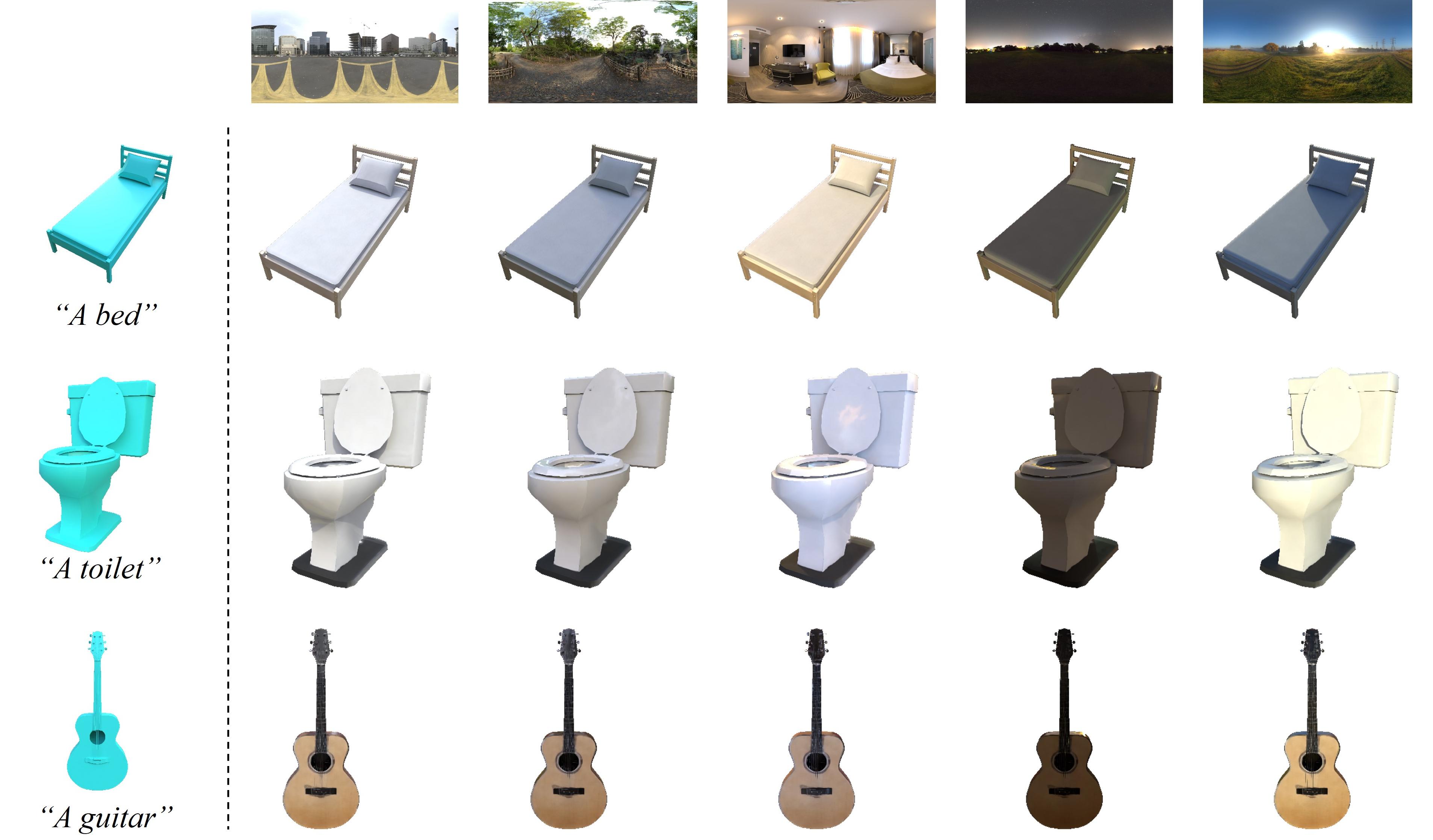}
    \caption{\textbf{Relighting results by Material Anything under various HDR environment maps.} The left column displays the input texture-less meshes, while the top row presents the HDR environment maps used. }
    \label{fig:relighting}
\end{figure*}

\section{Applications}
Material Anything offers robust capabilities to edit and customize materials of texture-less 3D objects by simply adjusting the input prompt, enabling flexible and intuitive material manipulation. As illustrated in Fig.~\ref{fig:editing}, we demonstrate that a barrel's material can be transformed into realistic textures like wood, gold, and stone, showcasing the versatility of our approach across various material types. This application allows users to dynamically adapt 3D models to specific aesthetic or functional requirements, enhancing asset adaptability for virtual environments, simulations, and design visualization. 

Furthermore, our method supports relighting, enabling objects to be viewed under different lighting conditions, as shown in Fig.~\ref{fig:relighting}. Material Anything generates material properties for each object, ensuring physically consistent relighting and enhanced realism. This functionality allows for more accurate simulations in AR, VR, and digital content creation, where realistic lighting is essential for immersion. Collectively, these capabilities make the proposed method a versatile and efficient solution for content creators and researchers aiming to produce high-quality, relightable 3D objects with customized materials.
\begin{figure}[t]
    \centering
    \includegraphics[width=\linewidth]{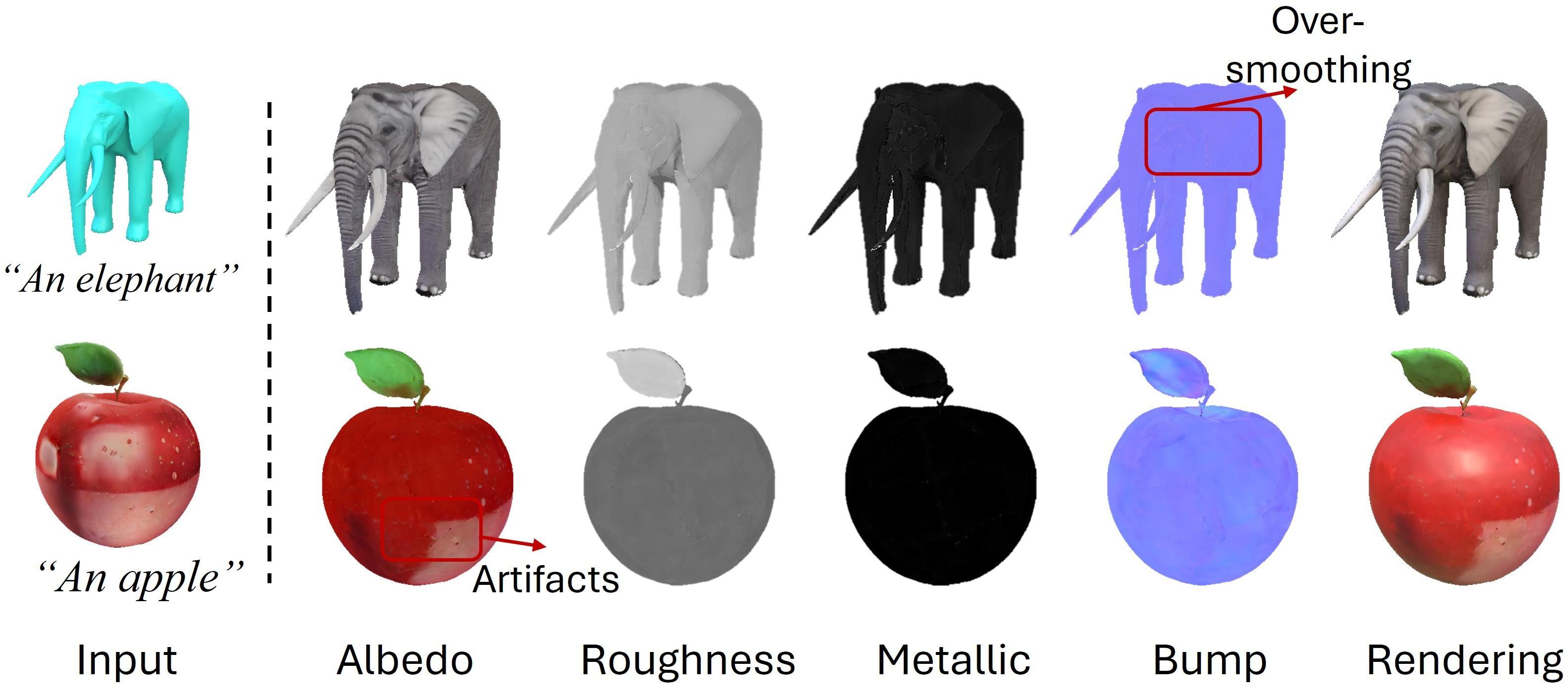}
    \caption{\textbf{Failure Cases by Material Anything.}}
    \label{fig:limitations}
\end{figure}
\section{Limitations and Failure Cases}
Material Anything is designed to address the complex task of generating materials for a diverse range of 3D objects. However, our approach has certain limitations. First, owing to the characteristics of the Objaverse, where many objects exhibit uniform roughness and metallic attributes with minimal surface details in bump maps, our method may produce materials with constrained surface details. This limitation is illustrated in the elephant example in Fig.~\ref{fig:limitations}, where the resulting bump maps lack details. Additionally, for objects with existing textures, our method struggles to remove prominent artifacts. For example, in the apple instance in Fig.~\ref{fig:limitations}, large white artifacts are misinterpreted as part of the texture, resulting in an inaccurate albedo.

\section{Additional Results}
\begin{figure*}
    \centering
    \includegraphics[width=0.82\textwidth]{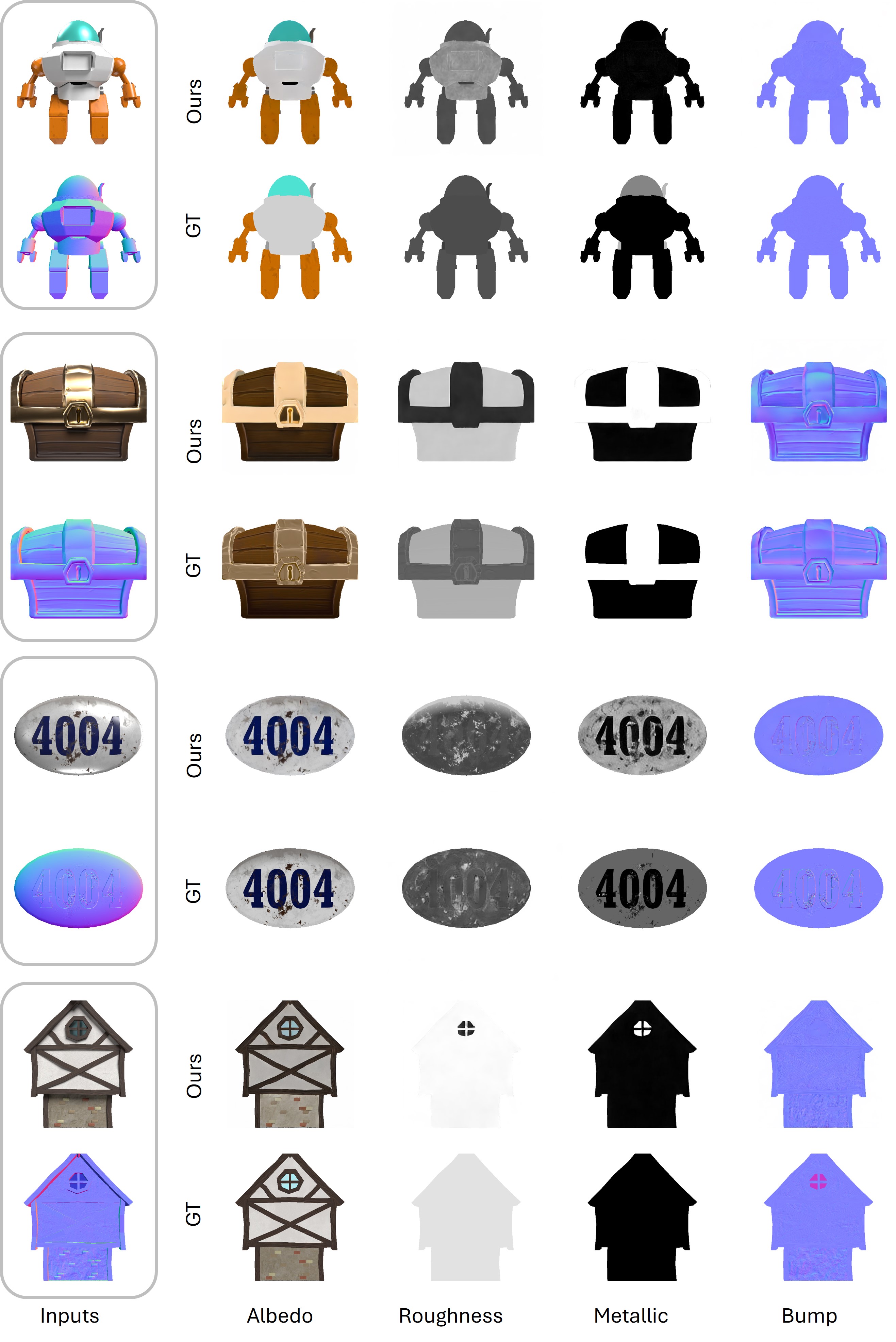}
    \caption{\textbf{Results by our material estimator on 2D renderings from Objaverse.}}
    \label{fig:2D_results}
\end{figure*}

\begin{figure*}
    \centering
    \includegraphics[width=0.85\textwidth]{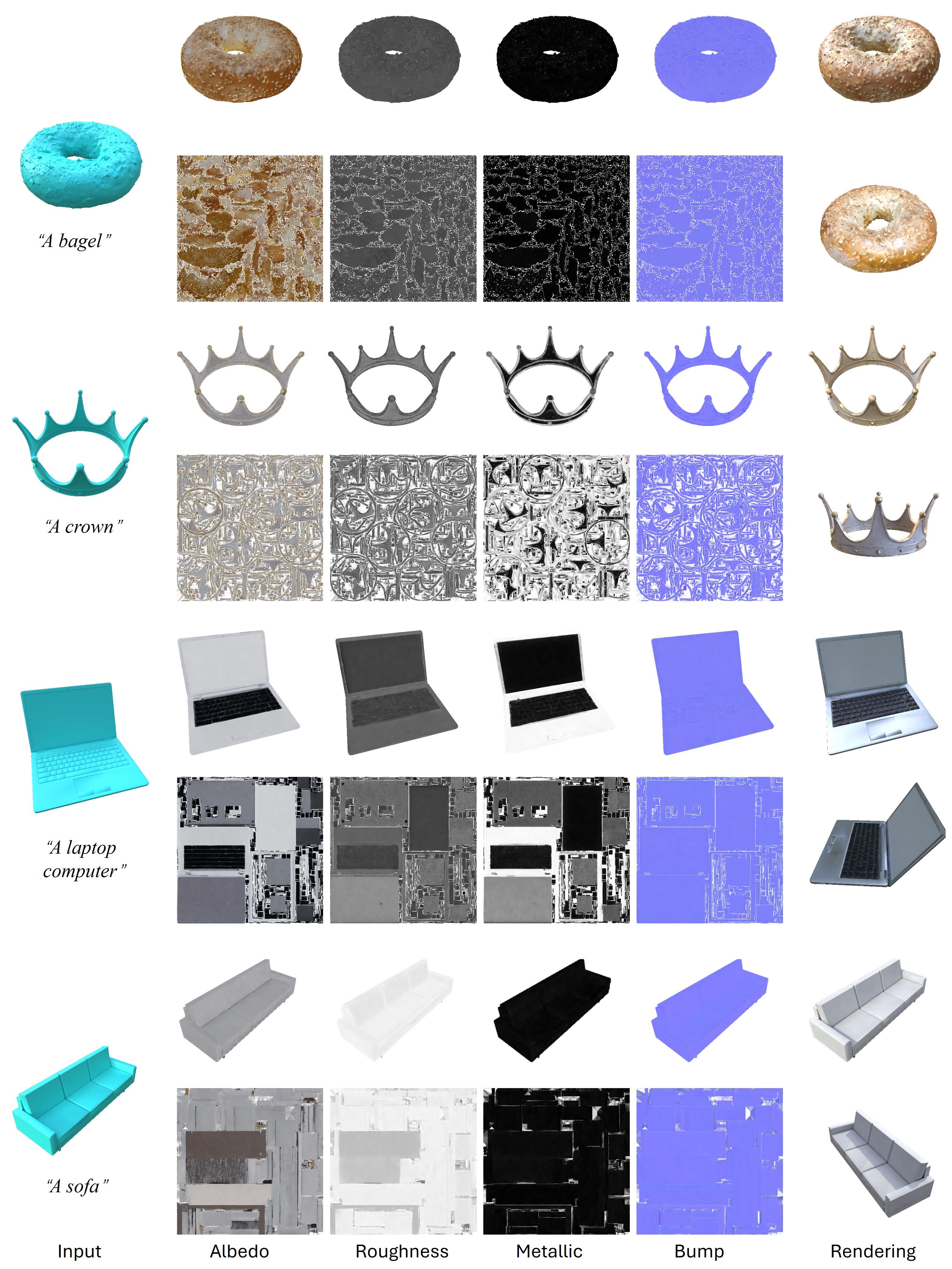}
    \caption{\textbf{Additional results by Material Anything on texture-less 3D objects.} The generated UV material maps are provided.}
    \label{fig:more_texture_less_objects}
\end{figure*}

\begin{figure*}
    \centering
    \includegraphics[width=0.90\textwidth]{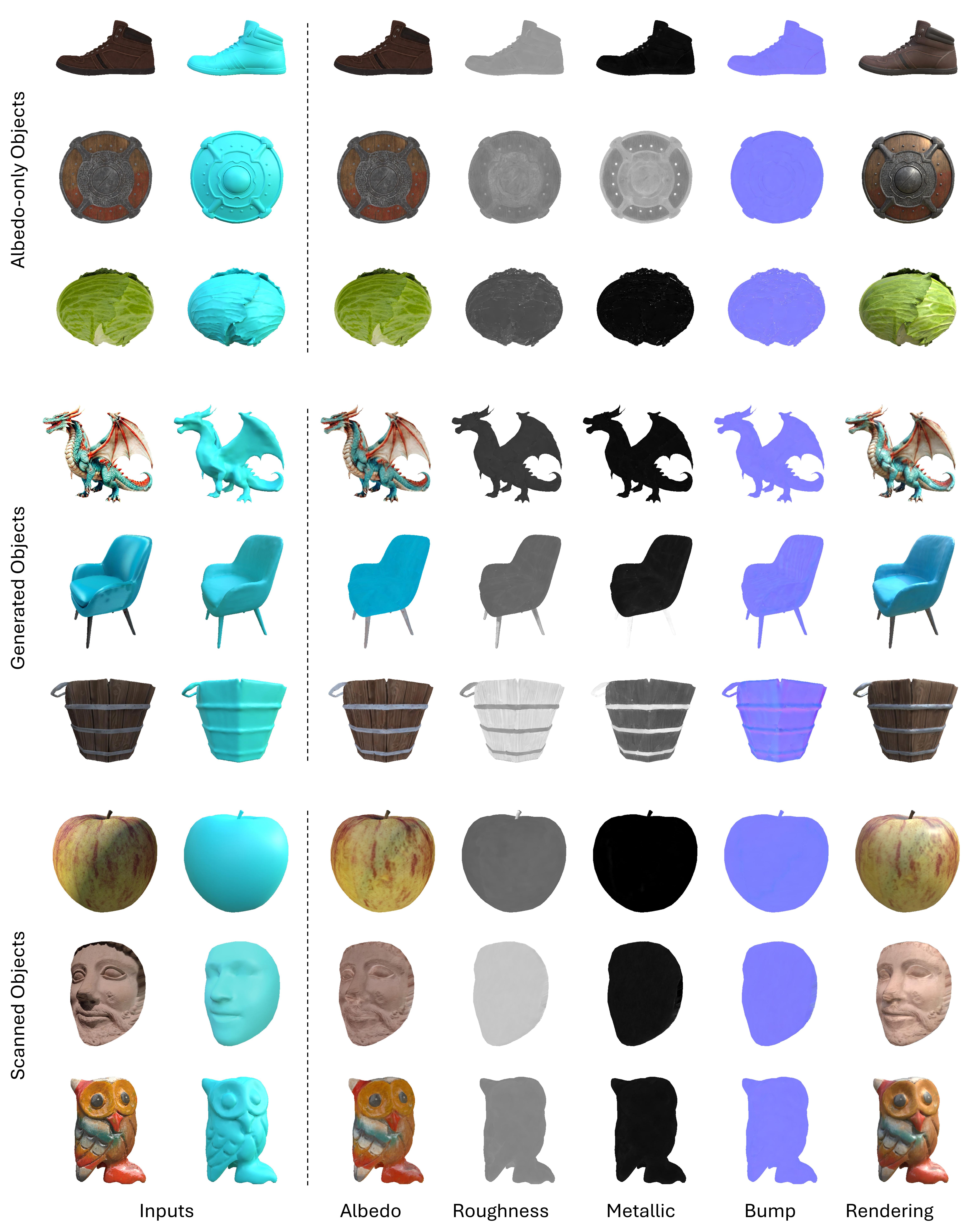}
    \caption{\textbf{Additional results by Material Anything on albedo-only, generated, scanned 3D objects.}}
    \label{fig:more_texture_objects}
\end{figure*}

We present additional qualitative results to illustrate the effectiveness of Material Anything. \textbf{Video results are present in our supplementary video.} In Fig.~\ref{fig:2D_results}, we display results generated by our material estimator on the Objaverse dataset, compared with their GT materials. As shown, our method effectively generates materials closely aligned with the ground truth, capturing essential details and textures to enhance realism. In Fig.~\ref{fig:more_texture_less_objects}, we show additional results on texture-less inputs, demonstrating our method’s capability to handle complex UV mappings. Despite the complexity of certain UV layouts, our method consistently generates high-quality material maps in UV space, preserving material fidelity across the entire surface. Finally, we present additional results on various input types, including generated models, albedo-only inputs, and scanned 3D objects. These examples, shown in Fig.~\ref{fig:more_texture_objects}, highlight our method’s robustness across varied lighting conditions and input characteristics, demonstrating its versatility in producing realistic materials adaptable to diverse lighting environments.

